\documentclass[11pt]{article}

\usepackage[margin=1in]{geometry}

\usepackage[T1]{fontenc}
\usepackage[utf8]{inputenc}
\usepackage{lmodern}
\usepackage{microtype}

\usepackage{amsmath,amssymb,amsthm}
\usepackage{pifont}
\newcommand{\cmark}{\ding{51}}
\newcommand{\xmark}{\ding{55}}
\usepackage{enumitem}

\usepackage{graphicx}
\usepackage{booktabs}
\usepackage{multirow}
\usepackage{tabularx}
\usepackage{caption}
\usepackage{subcaption}
\usepackage[table]{xcolor}

\definecolor{tableheader}{gray}{0.92}

\usepackage{placeins}


\usepackage{tcolorbox}
\tcbuselibrary{breakable,skins}

\usepackage{listings}
\usepackage[colorlinks=true,linkcolor=blue!70!black,citecolor=blue!70!black,
  urlcolor=blue!70!black,hyperfootnotes=false]{hyperref}

\usepackage{natbib}
\newcommand{\norm}[1]{\left\lVert #1 \right\rVert}

\newcommand{\bmat}[1]{\mathbf{#1}}
\newcommand{\reals}{\mathbb{R}}
\newcommand{\bigO}{\mathcal{O}}

\makeatletter
\@ifundefined{iflatexml}{\newif\iflatexml\latexmlfalse}{}
\makeatother

\newcommand{\tms}{\texttimes}

\newenvironment{papercallout}[1]{%
  \iflatexml
    \begin{center}\small
    \setlength{\tabcolsep}{6pt}%
    \begin{tabularx}{0.97\linewidth}{|>{\raggedright\arraybackslash}X|}
      \hline
      \rowcolor{tableheader}\textbf{#1}\\\hline
      \begin{minipage}[t]{0.93\linewidth}\vspace{4pt}\small
  \else
    \begin{tcolorbox}[title={\textbf{#1}},colback=white,colframe=gray!50,
      fonttitle=\small\bfseries,left=6pt,right=6pt,top=4pt,bottom=4pt]
    \small
  \fi
}{%
  \iflatexml
      \vspace{4pt}\end{minipage}\\\hline
    \end{tabularx}
    \end{center}
  \else
    \end{tcolorbox}
  \fi
}

\title{Scaling DoRA: High-Rank Adaptation via Factored Norms and
  Fused Kernels}
\author{
  Alexandra Zelenin\thanks{Correspondence: \texttt{alexa@eyes.ml}}
  \and
  Alexandra Zhuravlyova
}

\begin{document}
\maketitle

\begin{abstract}
Weight-Decomposed Low-Rank Adaptation (DoRA;~\cite{liu2024dora}) extends LoRA
by decoupling weight magnitude from direction, but its forward pass requires the
row-wise norm $\norm{\bmat{W} + s\bmat{B}\bmat{A}}_\text{row}$, a computation
that every major framework we surveyed implements by materializing the dense
$[d_\text{out}, d_\text{in}]$ product $\bmat{B}\bmat{A}$.
At $d_\text{in}=8192$ and rank $r=384$, a single module's norm requires
${\sim}512$\,MB of transient working memory in bf16, making high-rank
DoRA costly and often infeasible on common single-GPU setups once hundreds of
adapted modules and checkpointing are involved.

We present two systems contributions: a \emph{factored norm} that decomposes the
squared norm into base, cross, and Gram terms computable through
$\bigO(d_\text{out} r + r^2)$ intermediates, eliminating the dense product.
\emph{Fused Triton kernels} collapse the four-kernel DoRA composition into a
single pass, reducing memory traffic by ${\sim}$4\tms{} and using a numerically
stable form that avoids catastrophic cancellation in the near-unity
rescaling regime where magnitude scales concentrate in practice.

Across six 8--32B vision-language models (VLMs) on three NVIDIA GPUs
 (RTX~6000~PRO, H200, B200)
at $r\!=\!384$ in bf16, the fused implementation is $1.5$--$2.0$\tms{} faster
than HF~PEFT's DoRA implementation for inference, and $1.5$--$1.9$\tms{} faster
for gradient computation (optimizer step excluded), with up to
$7$\,GB lower peak VRAM.  Microbenchmarks on six GPUs
spanning four architecture generations (L40S, A100, RTX~6000~PRO, H200, B200,
B300) confirm $1.5$--$2.7$\tms{} compose-kernel speedup.  Final-logit cosine
similarity exceeds $0.9999$ across all model/GPU pairs, and multi-seed training
curves match within $7.1 \times 10^{-4}$ mean per-step loss delta over 2000 steps.
\end{abstract}

\section{Introduction}
\label{sec:intro}

Low-Rank Adaptation (LoRA;~\citealt{hu2022lora}) is the dominant method for
parameter-efficient fine-tuning.  DoRA~\citep{liu2024dora} extends LoRA by
decomposing the adapted weight into magnitude and direction:
\begin{equation}
  \bmat{W}' = \bmat{m} \odot \frac{\bmat{W} + s\bmat{B}\bmat{A}}
  {\norm{\bmat{W} + s\bmat{B}\bmat{A}}_\text{row}}
  \label{eq:dora}
\end{equation}
where $\bmat{W} \in \reals^{d_\text{out} \times d_\text{in}}$ is the frozen
base weight, $\bmat{B} \in \reals^{d_\text{out} \times r}$ and
$\bmat{A} \in \reals^{r \times d_\text{in}}$ are low-rank factors, $s$ is a
scaling coefficient (e.g., rsLoRA;~\citealt{kalajdzievski2023rslora}), and
$\bmat{m} \in \reals^{d_\text{out}}$ is a learnable magnitude vector.
High-rank configurations narrow the gap to full fine-tuning on complex
downstream tasks~\citep{hu2022lora,liu2024dora}.  We treat weights as
$[d_\text{out}, d_\text{in}]$ and compute per-output-row norms
(\texttt{dim=1}), consistent with PEFT and torchtune.

The bottleneck is the row-wise $L_2$ norm of the composed weight
$\bmat{W} + s\bmat{B}\bmat{A}$.
Hugging Face PEFT~\citep{peft} (and five other major frameworks we
surveyed: torchtune, Unsloth, SWIFT, LLaMA-Factory, Axolotl; see
Appendix~\ref{app:framework-survey}) computes this by constructing a
$d_\text{in} \times d_\text{in}$ identity matrix, thereby materializing the dense
product $\bmat{B}\bmat{A}$:
\begin{lstlisting}
x_eye = torch.eye(lora_A.weight.shape[1], ...)  # [d_in, d_in]
lora_weight = lora_B(lora_A(x_eye)).T            # [d_out, d_in]
weight_norm = torch.linalg.norm(weight + scaling * lora_weight, dim=1)
\end{lstlisting}
This incurs $\bigO(d_\text{in}^2)$ memory for the identity matrix alone:
32\,MB at $d_\text{in}=4096$, 128\,MB at $d_\text{in}=8192$ in bf16.
Including the dense $\bmat{B}\bmat{A}$ product and composed-weight copy, a
single module allocates 3--4 dense $[d_\text{out}, d_\text{in}]$ temporaries:
${\sim}512$\,MB at $d_\text{in}=8192$.  With gradient
checkpointing~\citep{chen2016checkpointing}, these temporaries are allocated
\emph{twice} per step.  Across hundreds of adapted modules in an 8--32B model,
this cumulative pressure is a major contributor to both speed degradation and
OOM failures at high rank.

The most obvious fix (computing \texttt{lora\_B.weight @ lora\_A.weight}
directly) eliminates the identity matrix but still materializes the full
$[d_\text{out}, d_\text{in}]$ product, which is the dominant cost.  We show
in \S\ref{sec:dense-ba} that this ``dense (B@A)'' path provides inconsistent
speedups that depend on GPU bandwidth class and sometimes runs \emph{slower}
than the PEFT baseline.

This paper does not propose a new adapter architecture, optimizer, or training
recipe.  Our contribution is systems-oriented: we execute the same DoRA
computation with a smaller working set and lower memory traffic.
Specifically:
\begin{enumerate}
  \item A \textbf{factored norm computation} (\S\ref{sec:factored-norm})
    decomposes $\norm{\bmat{W} + s\bmat{B}\bmat{A}}^2_\text{row}$ into three
    terms, each evaluable through $\bigO(d_\text{out} r + r^2)$ intermediates
    without materializing $\bmat{B}\bmat{A}$.  At $d=8192$, $r=512$ in fp32,
    the theoretical persistent-memory reduction is $15$\tms{}
    (Table~\ref{tab:complexity}).

  \item \textbf{Fused Triton kernels} (\S\ref{sec:triton}) collapse the
    DoRA composition $(g{-}1)\!\odot\!\text{base} +
    g\!\odot\!s\!\odot\!\text{lora}$ from four CUDA kernel launches to one
    pass.  A numerically stable form avoids catastrophic cancellation when
    $g \approx 1$.  Forward speedup: $1.5$--$2.7$\tms{} (geometric mean);
    backward speedup: $1.06$--$1.23$\tms{}.  A three-tier runtime dispatch
    (\S\ref{sec:dispatch}) selects the optimal path (fused backward for
    training, fused forward for inference, eager fallback for CPU or
    sub-crossover shapes), compatible with \texttt{torch.compile}~\citep{ansel2024pytorch2},
    gradient checkpointing, DeepSpeed ZeRO~\citep{rajbhandari2020zero}, and FSDP1.
\end{enumerate}
Both contributions are validated on six NVIDIA GPUs spanning four architecture
generations (L40S, A100, RTX~6000~PRO, H200, B200, B300; 48--268\,GB) with
model-level benchmarks on three GPUs across six 8--32B VLMs
(\S\ref{sec:experiments}).  Throughout this paper, four configurations are
compared: \emph{PEFT} (unmodified HF~PEFT identity-matrix path),
\emph{Dense~(B@A)} (direct product, still materializes the full matrix),
\emph{Eager} (our factored norm with PyTorch composition), and \emph{Fused}
(our factored norm with Triton kernels).

\section{Factored Norm Computation}
\label{sec:factored-norm}

\subsection{Algebraic Decomposition}

The row-wise squared norm of the composed weight expands into three terms:
\begin{equation}
  \norm{\bmat{W} + s\bmat{B}\bmat{A}}_\text{row}^2
  = \underbrace{\norm{\bmat{W}}_\text{row}^2}_{\text{base}}
  + \underbrace{2s \langle \bmat{W}, \bmat{B}\bmat{A}
    \rangle_\text{row}}_{\text{cross}}
  + \underbrace{s^2 \norm{\bmat{B}\bmat{A}}_\text{row}^2}_{\text{BA norm}}
  \label{eq:factored-norm}
\end{equation}
where $\langle \cdot, \cdot \rangle_\text{row}$ denotes the row-wise inner
product.  Each term is computable through low-rank intermediates:

\paragraph{Base norm.}
$\norm{\bmat{W}}_\text{row}^2$ accumulates via chunks along $d_\text{in}$,
producing a vector of size $d_\text{out}$.  Chunking limits working memory to a
configurable budget (default: 256\,MB).

\paragraph{Cross term.}
The row-wise inner product rewrites as:
\begin{equation}
  \langle \bmat{W}, \bmat{B}\bmat{A} \rangle_j
  = \sum_\ell B_{j\ell} \cdot U_{j\ell}
  = (\bmat{B} \odot \bmat{U})_j \cdot \mathbf{1}
  \label{eq:cross-term}
\end{equation}
where $\bmat{U} = \bmat{W}\bmat{A}^\top \in \reals^{d_\text{out} \times r}$
accumulates chunk-wise:
$\bmat{U} \leftarrow \bmat{U} + \bmat{W}_c \bmat{A}_c^\top$.

\paragraph{BA norm.}
The row-wise squared norm factors through the Gram matrix:
\begin{equation}
  \norm{\bmat{B}\bmat{A}}_j^2
  = (\bmat{B} \bmat{G} \odot \bmat{B})_j \cdot \mathbf{1}
  \label{eq:ba-norm}
\end{equation}
where $\bmat{G} = \bmat{A}\bmat{A}^\top \in \reals^{r \times r}$ also
accumulates chunk-wise.  At $r=512$ in fp32, $\bmat{G}$ occupies 1\,MB.

\subsection{Assembly and Precision}

The three per-row scalars assemble into the weight norm:
\begin{equation}
  w_\text{norm} = \sqrt{\max\!\left(
    \norm{\bmat{W}}_\text{row}^2
    + 2s \cdot \text{cross}
    + s^2 \cdot \text{ba\_norm},\; 0
  \right)}
  \label{eq:norm-assembly}
\end{equation}
The magnitude division is always computed in PyTorch after the kernel returns:
\begin{equation}
  g \triangleq \bmat{m} \,/\, \max(w_\text{norm}, \epsilon)
  \label{eq:mag-scale}
\end{equation}
This ensures identical precision regardless of whether the Triton or PyTorch
norm path produced $w_\text{norm}$, eliminating a source of fidelity divergence
we observed at large activation scales (see \S\ref{sec:model-perf}).

All accumulation is performed in fp32 under \texttt{torch.no\_grad()} with
\texttt{autocast} disabled.  Disabling autocast alone does not force fp32 when
inputs are bf16, so each chunk of $\bmat{W}$, $\bmat{A}$, $\bmat{B}$, and the
intermediate $\bmat{U}_c$ is explicitly cast to fp32 before accumulation.
This is consistent with the DoRA paper's instruction (Section~4.3) to treat the
norm as a detached constant~\citep{liu2024dora}.  We use $g$ consistently
throughout to denote the post-division scale, distinct from the learnable
magnitude $\bmat{m}$.

\subsection{Complexity}

Table~\ref{tab:complexity} compares asymptotic and concrete memory costs.

\begin{table}[t]
  \centering
  \caption{The factored norm reduces rank-dependent persistent memory by
    $15$\tms{} at $d\!=\!8192$, $r\!=\!512$ in fp32.  Measured reductions are
    smaller ($3.2$\tms{}) because allocator deltas include the rank-independent
    base-norm transient (\S\ref{sec:complexity-detail}).}
  \label{tab:complexity}
  \small
  \begin{tabular}{lcc}
    \toprule
    \rowcolor{tableheader} & \textbf{PEFT} & \textbf{Factored (ours)} \\
    \midrule
    Rank-dependent persistent & $\bigO(d_\text{in}^2 + d_\text{out} d_\text{in})$
      & $\bigO(d_\text{out} r + r^2)$ \\
    Chunk buffer (transient) & Included above
      & $[d_\text{out}, \text{cs}]$ per chunk$^\dagger$ \\
    Per-module temps & 3--4 dense $[d_\text{out}, d_\text{in}]$
      & $\bmat{U}\![d_\text{out},r]$ + $\bmat{G}\![r,r]$ \\
    Identity matrix & Every forward call & Never created \\
    \midrule
    \multicolumn{3}{l}{\emph{Concrete}: $d_\text{out}\!=\!d_\text{in}\!=\!8192$,
      $r\!=\!512$, fp32} \\
    \midrule
    Theory: dense (B@A) & 256\,MB & N/A \\
    Theory: U + G & N/A & 17.0\,MB \\
    \textbf{Theoretical reduction} & \multicolumn{2}{c}{$\mathbf{15.1}$\tms{}} \\
    \midrule
    Measured: allocator delta & 768\,MB & 241\,MB \\
    \textbf{Measured reduction} & \multicolumn{2}{c}{$\mathbf{3.2}$\tms{}} \\
    \bottomrule
    \multicolumn{3}{l}{\footnotesize $\dagger$\,$\text{cs} =
      \min(d_\text{in}, \lfloor\text{budget}/(d_\text{out} \cdot 4)\rfloor)$;
      at 256\,MB and $d{=}8192$, cs spans full $d_\text{in}$.} \\
  \end{tabular}
\end{table}

\paragraph{Why the measured reduction is smaller.}
\label{sec:complexity-detail}
The dominant transient is the base-norm computation (Term~1 of
Equation~\ref{eq:factored-norm}): the chunked
$\norm{\bmat{W}}_\text{row}^2$ accumulation creates a
$[d_\text{out}, \text{chunk\_size}]$ fp32 buffer that, at the default budget
and $d=8192$, approaches 256\,MB, accounting for most of the 241\,MB measured
delta.  This cost is \emph{rank-independent}: identical at $r=16$ and $r=768$.
The theoretical reduction, which counts only rank-dependent tensors ($\bmat{U}$
and $\bmat{G}$), correctly predicts the asymptotic benefit as rank grows.

Since $\bmat{W}$ is frozen, $\norm{\bmat{W}}^2_\text{row}$ could be
precomputed into a $[d_\text{out}]$ buffer (16\,KB at $d_\text{out}=4096$),
eliminating this transient entirely.  We leave this caching for future work.

\paragraph{bf16 caveat.}
The factored norm accumulates in fp32 regardless of weight dtype.  Against
half-precision PEFT baselines, this fp32 overhead inverts the isolated-norm
memory ratio (PEFT/factored) to $0.8$\tms{} (i.e., factored uses \emph{more} for the norm
micro-operation in bf16).  This does not negate model-level VRAM savings
(Table~\ref{tab:model-memory-comparison}), which include the fused compose
kernel's elimination of forward-pass intermediates.

\paragraph{Compute tradeoff.}
The factored norm is ${\sim}$4.8\tms{} slower than the dense reference when measured
isolation (H200, fp32) because the reference performs a single contiguous
\texttt{torch.linalg.norm} call, while the factored path uses multiple
chunked matmuls.  The system is faster end-to-end because the reference
\emph{first} materializes the full $[d_\text{out}, d_\text{in}]$ product;
it is this materialization, not the norm itself, that dominates time and memory.
On lower-bandwidth hardware (RTX~6000~PRO, GDDR7), the factored norm matches
or outperforms the reference at production ranks ($r \leq 384$) for large weight
matrices, so the $4.8$\tms{} figure is a conservative bound.

\begin{papercallout}{Algorithm 1: Factored Row-wise Norm}
\label{alg:factored-norm}
\textbf{Input:} $\bmat{W} \in \mathbb{R}^{d_\text{out} \times d_\text{in}}$
(frozen); $\bmat{A} \in \mathbb{R}^{r \times d_\text{in}}$,
$\bmat{B} \in \mathbb{R}^{d_\text{out} \times r}$;
$s \in \mathbb{R}$; $\varepsilon$ (dtype-dependent);
\texttt{chunk\_budget} (bytes, default 256\,MB)

\medskip
$\text{cs} \leftarrow \min(d_\text{in},\,
  \lfloor\text{budget}/(d_\text{out}\cdot 4)\rfloor)$,
aligned to 64 elements.

\smallskip
\textit{All accumulation in fp32 under \texttt{torch.no\_grad()}, autocast
disabled.  Cast $\bmat{W}$ chunks, $\bmat{A}$, $\bmat{B}$ to fp32 on the fly.}

\smallskip
Initialize: $\text{base\_sq} \leftarrow \mathbf{0}_{d_\text{out}}$,\;
$\text{cross} \leftarrow \mathbf{0}_{d_\text{out}}$,\;
$\bmat{G} \leftarrow \mathbf{0}_{r \times r}$ \quad(all fp32)

\medskip
\textbf{for} each chunk $c$ of size cs: \\[1pt]
\quad \texttt{W\_c = W[:, c:c+cs].float()}
  \hfill $[d_\text{out}, \text{cs}]$ \\[1pt]
\quad \texttt{A\_c = A[:, c:c+cs].float()}
  \hfill $[r, \text{cs}]$ \\[1pt]
\quad \texttt{base\_sq += (W\_c**2).sum(dim=1)}
  \\[1pt]
\quad $\bmat{G} \leftarrow \bmat{G} + \bmat{A}_c \bmat{A}_c^\top$ \\[1pt]
\quad $\bmat{U}_c \leftarrow \bmat{W}_c \bmat{A}_c^\top$
  \hfill $[d_\text{out}, r]$\;\textit{(not retained)} \\[1pt]
\quad \texttt{cross += (B.float() * U\_c).sum(dim=1)}

\medskip
\texttt{ba\_sq = (B.float() @ G * B.float()).sum(dim=1)}
  \hfill $[d_\text{out}]$ \\[2pt]
$w_\text{norm} \leftarrow \sqrt{\max(\text{base\_sq}
  + 2s\cdot\text{cross} + s^2\cdot\text{ba\_sq},\; 0)}$
  \hfill $[d_\text{out}]$ \\[2pt]
\textbf{return} \texttt{w\_norm.to(input\_dtype)}

\medskip
\textbf{Notes:}
Chunk size aligns to 64 for Tensor Core MMA.
$\bmat{U}_c$ is never stored for multiple chunks simultaneously.
When $s\!=\!0$, cross and ba\_sq are skipped; $\bmat{U}_c$ and $\bmat{G}$ are
not allocated.
$\bmat{G}$ is $\leq 2.4$\,MB at $r=768$ in fp32.
\end{papercallout}

\section{Fused Triton Kernels}
\label{sec:triton}

\subsection{Compose Kernel}

The DoRA composition $(g{-}1) \odot \text{base} + g \odot s \odot \text{lora}$
decomposes into four sequential element-wise operations in standard PyTorch,
each launching a separate CUDA kernel: 3 reads + 1 write per op yields
${\sim}12$ memory passes total.  The fused
Triton~\citep{tillet2019triton} kernel collapses these into a single pass:
3 reads (base, lora, $g$) + 1 write, a ${\sim}$4\tms{} reduction in memory
traffic.  The realized speedup of $2.0$--$2.7$\tms{} (rather than $4$\tms{})
reflects the fact that the eager path is partially latency-bound by kernel-launch gaps;
the fused kernel achieves ${\sim}50\%$ of peak HBM bandwidth
(Figure~\ref{fig:bandwidth}), vs.\ ${\sim}20\%$ for the eager path.

\paragraph{Numerical stability.}
The algebraically equivalent form
$g \odot (s \cdot \text{lora} + \text{base}) - \text{base}$ suffers from
catastrophic cancellation when $g \approx 1$.  This regime is not hypothetical.
The stored magnitude parameters reflect the heterogeneous row norms of
pretrained weights and naturally vary across layers and models, but DoRA
initializes $\bmat{m} = \norm{\bmat{W}}_\text{row}$ and magnitudes track
weight norms throughout training, so the composed scale
$g = \bmat{m}/w_\text{norm}$ concentrates tightly around unity (mean
$\approx 1.0$, std $\approx 0.0015$).  Measurement on a Qwen2-VL-7B adapter
($r\!=\!128$, 326 modules, 1.77M elements) shows that 100\% of $g$ values fall
in the bf16 collapse zone ($|g-1| < \varepsilon_\text{bf16}/2$) and 20\% in
the fp16 zone: if $(g{-}1) \odot \text{base}$ were evaluated in bf16, the base
correction would vanish for every element; in fp16, for one in five.

\begin{figure}[t]
  \centering
  \includegraphics[width=0.88\linewidth]{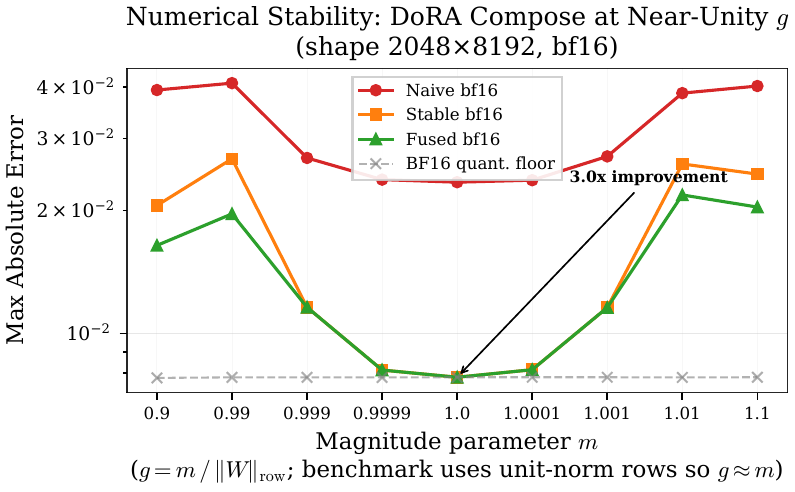}
  \caption{The stable compose form achieves $3.0$\tms{} lower peak error
    near $g \approx 1$ (bf16, $d_\text{out}=8192$, $d_\text{in}=2048$).
    The naive form $g \odot (s \cdot \text{lora} + \text{base}) - \text{base}$
    exhibits catastrophic cancellation; the stable form and fused kernel both
    remain near the bf16 quantization floor.  Reference: fp64.}
  \label{fig:stability}
\end{figure}

The stable form $(g{-}1) \odot \text{base} + g \odot s \odot \text{lora}$
keeps the small correction $(g{-}1)$ explicit, but its precision advantage
depends on fp32 intermediate computation to prevent $(g{-}1)$ from rounding to
zero.  Both the Triton kernel and PyTorch fallback use this form with fp32
compute.  Figure~\ref{fig:stability} shows $3.0$\tms{} lower peak error
near $g \approx 1$ compared to the naive alternative.  Beyond the algebraic
form, bf16 multiplication is non-associative: all code paths enforce a single
canonical evaluation order ($s \cdot \text{lora}$ first, then
$g \cdot (\cdot)$), ensuring bitwise parity across all PyTorch composition
paths.

\paragraph{Autotuning.}
Optimal kernel configurations vary substantially across GPUs (${\sim}9\%$
pairwise agreement across six GPUs), requiring per-device autotuning rather
than a static table.  First-run autotuning takes 10--30\,s per kernel, and caches
persist in Triton's default directory.  Details in
Appendix~\ref{app:autotune}.

\subsection{Backward Kernel}

The fused backward computes
$d_\text{lora} = g \cdot s \cdot d_\text{out}$ and
$d_\text{base} = (g{-}1) \cdot d_\text{out}$ in a single Triton pass.
Two design decisions merit note:
\begin{itemize}
  \item \textbf{Reduced \texttt{ROWS\_PER\_PROGRAM}}: Writing two output tensors
    doubles per-element traffic; reducing rows per program lowers register
    pressure and improves SM utilization.
  \item \textbf{$d_\text{mag}$ via PyTorch reduction}: The magnitude gradient
    uses a separate \texttt{.sum()} rather than \texttt{tl.atomic\_add},
    avoiding contention at large \texttt{num\_rows} and the
    non-deterministic ordering of floating-point atomics.
\end{itemize}

\subsection{Norm Assembly Kernel}

A second Triton kernel fuses Equation~\ref{eq:norm-assembly}, computing
$w_\text{norm}$ from the three factored terms.  Store-reload barriers prevent
FMA fusion, and an inline PTX \texttt{sqrt.rn.f32} instruction replaces
Triton's default approximate \texttt{sqrt}, exactly reproducing PyTorch's
evaluation order.  The kernel stops at $w_\text{norm}$; the magnitude
division (Equation~\ref{eq:mag-scale}) remains in PyTorch so both norm paths
share the same precision context.  Appendix~\ref{app:kernel-spec} provides
exact specifications for all three kernels.

\section{Runtime Dispatch}
\label{sec:dispatch}

\begin{figure}[t]
  \centering
  \includegraphics[width=0.92\linewidth]{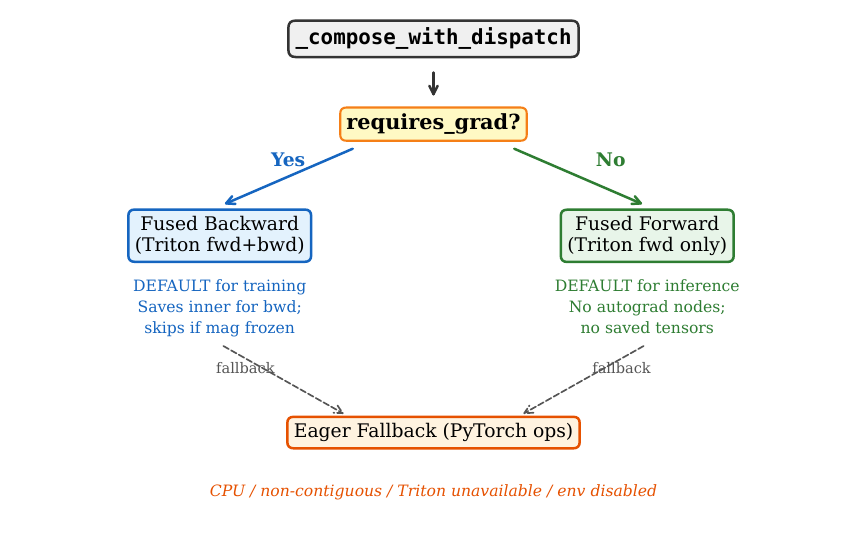}
  \caption{Three-tier dispatch: fused backward for training (Tier~1),
    fused forward for inference (Tier~2), eager fallback for CPU,
    no-Triton, or sub-crossover shapes (Tier~3).}
  \label{fig:dispatch}
\end{figure}

The composition path is selected at runtime by
\texttt{\_compose\_with\_dispatch} (Figure~\ref{fig:dispatch},
Table~\ref{tab:dispatch-tiers}).  Four environment variables control kernel
availability and working-set budgets; defaults require no configuration.

\begin{table}[t]
  \centering
  \small
  \caption{Dispatch tiers and their selection criteria.}
  \label{tab:dispatch-tiers}
  \begin{tabularx}{\linewidth}{@{}ll>{\raggedright\arraybackslash}X
    >{\raggedright\arraybackslash}X@{}}
    \toprule
    \rowcolor{tableheader} \textbf{Tier} & \textbf{Path} & \textbf{When}
      & \textbf{Tradeoff} \\
    \midrule
    1 & Fused Backward & Training + CUDA + Triton + auto-gate/force-on
      & Fastest above crossover; saves \texttt{inner} for bwd \\
    2 & Fused Forward  & Inference + CUDA + Triton
      & Speed without bwd memory \\
    3 & Eager Fallback & CPU / no Triton / force-off / sub-crossover
      & Universal compatibility \\
    \bottomrule
  \end{tabularx}
\end{table}

\paragraph{Tier~1 (Fused Backward).}
A dual-output Triton kernel computes both the output and the saved tensor
$\text{inner} = s \cdot \text{lora} + \text{base}$ in a single pass,
eliminating the forward-pass VRAM spike from sequential PyTorch ops.
When the magnitude is frozen (\texttt{requires\_grad=False}), the
\texttt{inner} allocation is skipped entirely.
The default auto-mode crossover requires
$d_\text{out} \geq 2048$ and
$(\text{batch} \times \text{seq}) \times d_\text{out} \geq 2048 \times 6144$;
smaller activations use Tier~3 because launch latency dominates.
In the six evaluated VLMs, KV projections ($d_\text{out}$ as low as 512) fall
below the crossover, so ${\sim}71\%$ of adapted modules per layer dispatch to
Tier~1 during training and ${\sim}29\%$ fall back to Tier~3.

\paragraph{Tier~2 (Fused Forward).}
A forward-only Triton kernel with no autograd graph nodes, dispatched when
\texttt{requires\_grad} is false.

\paragraph{Tier~3 (Eager Fallback).}
Pure PyTorch; handles CPU, no-Triton, and sub-crossover training.  Uses
out-of-place composition when autograd is active to avoid aliasing.

\paragraph{Precision.}
All PyTorch compose paths produce bitwise-identical forward outputs by
enforcing a single evaluation order.  The Triton kernels preserve the same
algebra but not bitwise equality (FMA contraction and reduction trees can
perturb last bits); we treat Triton--PyTorch agreement as an empirical
envelope: fp32 outputs stay within $10^{-4}$ max-abs error, bf16/fp16 remain
within dtype-appropriate tolerances (\S\ref{sec:model-perf}).

\paragraph{Compatibility.}
The fused compose is registered as a custom op
(\texttt{peft::\allowbreak{}fused\_dora\_compose}) via
\texttt{torch.library}, making the dispatch graph-break-free under
\texttt{torch.compile} when dropout is inactive ($p\!=\!0$).
DeepSpeed ZeRO-2/3 and FSDP1 are supported; FSDP2/DTensor is not
(\S\ref{sec:discussion}).  The forward contract,
\texttt{torch.compile} details, and the chunked-dropout path are
specified in Appendices~\ref{app:forward-contract}
and~\ref{app:impl}.

\paragraph{Magnitude division.}
Across all tiers,
$g = \bmat{m} / \max(w_\text{norm}, \epsilon)$ is computed in PyTorch outside
the \texttt{no\_grad} norm context, ensuring identical precision regardless of
execution tier.

\section{Experiments}
\label{sec:experiments}

\subsection{Setup}

Microbenchmarks use six GPUs spanning four architecture generations
(Table~\ref{tab:hardware}); model-level benchmarks use three GPUs
(RTX~6000~PRO, H200, B200) with sufficient VRAM for the tested models.
All GPUs run identical software: PyTorch~2.10.0+cu130, Triton~3.6.0,
Transformers~5.2.0, CUDA~13.1, driver~580.126.09.  The PEFT baseline is
upstream commit \texttt{20a9829}
(\texttt{v0.18.0.rc0}).\footnote{Later HEAD \texttt{9cf86c7}
(2026-02-24) is algorithmically identical for training; see
\S\ref{sec:related}.}
Model-level benchmarks exclude the optimizer step to isolate DoRA
overhead and use a partial-sequence loss (1024 loss tokens) to match
production RLHF/GRPO memory profiles; full-sequence loss creates a
6--12\,GB logit spike that masks adapter working-set differences.
A sensitivity check at 4096 loss tokens confirms speedups are unchanged.
Each microbenchmark reports the median of 200 CUDA-event-timed trials
(10 warmup); model-level benchmarks use 20 repeats (3 warmup,
CV~$<$~1.7\%).
Memory measurement methodology and full reproducibility instructions are provided
in Appendix~\ref{app:repro}.

\begin{table}[t]
  \centering
  \caption{Benchmark hardware.  ``Micro'': microbenchmark coverage.
    ``Model'': full model-level gradient-computation and inference benchmarks.}
  \label{tab:hardware}
  \small
  \begin{tabular}{lcccc}
    \toprule
    \rowcolor{tableheader} \textbf{GPU} & \textbf{Arch} & \textbf{Memory}
      & \textbf{BW (TB/s)} & \textbf{Scope} \\
    \midrule
    L40S          & SM89 Ada       & 48\,GB GDDR6   & 0.86  & Micro \\
    A100-SXM4     & SM80 Ampere    & 80\,GB HBM2e   & 2.04  & Micro \\
    RTX 6000 PRO  & SM120 Blackwell & 96\,GB GDDR7  & 1.60  & Micro+Model \\
    H200          & SM90 Hopper    & 141\,GB HBM3e  & 4.80  & Micro+Model \\
    B200          & SM100 Blackwell & 192\,GB HBM3e & 7.70  & Micro+Model \\
    B300          & SM103 Blackwell & 268\,GB HBM3e & 7.70  & Micro \\
    \bottomrule
  \end{tabular}
\end{table}

\subsection{Model-Level Performance}

Table~\ref{tab:model-perf-comparison} summarizes the headline result:
gradient-computation speedup across six 8--32B VLMs on three GPUs.
The fused implementation is $1.46$--$1.87$\tms{} faster than HF~PEFT's DoRA
implementation and
$1.18$--$1.24$\tms{} faster than our own eager baseline, with
$1.3$--$6.7$\,GB lower peak VRAM (Table~\ref{tab:model-memory-comparison}).
These timings cover forward+backward only (excluding optimizer updates), so
the end-to-end wall-clock gain is smaller: in the 2000-step convergence run,
the same optimization reduced total training time by 8.3\% once optimizer,
data loading, and framework overhead were included
(\S\ref{sec:convergence-equivalence}).
The 32B models exceed the 96\,GB RTX~6000~PRO under \emph{all} configurations;
this is a capacity limit, not a method-specific regression.

\begin{table}[!tbp]
  \centering
  \caption{Gradient-computation speedup on 8--32B VLMs ($r\!=\!384$, bf16,
    seq=4096, bs=1, ga=8, \texttt{loss\_tokens=1024}, 20 repeats).
    The HF~PEFT DoRA baseline takes 46--87\% longer per iteration than fused.
    32B models OOM on RTX~6000~PRO (96\,GB) under all configurations.
    See Table~\ref{tab:model-abs-times} for absolute times.}
  \label{tab:model-perf-comparison}
  \footnotesize
  \begin{tabular}{lccc|ccc}
    \toprule
    \rowcolor{tableheader} & \multicolumn{3}{c|}{\textbf{Speedup vs.\ PEFT DoRA}}
      & \multicolumn{3}{c}{\textbf{Speedup vs.\ Eager}} \\
    \rowcolor{tableheader} \textbf{Model}
      & \textbf{RTX} & \textbf{H200} & \textbf{B200}
      & \textbf{RTX} & \textbf{H200} & \textbf{B200} \\
    \midrule
    Qwen2.5-VL-32B & OOM  & 1.73\tms{} & 1.74\tms{}
                   & OOM  & 1.20\tms{} & 1.22\tms{} \\
    Qwen3-VL-32B   & OOM  & 1.66\tms{} & 1.67\tms{}
                   & OOM  & 1.18\tms{} & 1.21\tms{} \\
    Qwen3.5-27B    & 1.51\tms{} & 1.57\tms{} & 1.57\tms{}
                   & 1.22\tms{} & 1.21\tms{} & 1.23\tms{} \\
    Gemma3-27B     & 1.53\tms{} & 1.61\tms{} & 1.56\tms{}
                   & 1.20\tms{} & 1.19\tms{} & 1.21\tms{} \\
    Mistral-Sm-24B & 1.66\tms{} & 1.87\tms{} & 1.87\tms{}
                   & 1.22\tms{} & 1.21\tms{} & 1.23\tms{} \\
    Qwen3-VL-8B   & 1.46\tms{} & 1.50\tms{} & 1.47\tms{}
                   & 1.23\tms{} & 1.21\tms{} & 1.24\tms{} \\
    \bottomrule
  \end{tabular}
\end{table}

\begin{table}[!tbp]
  \centering
  \caption{Absolute gradient-computation time (seconds).  Each iteration
    covers 8 gradient-accumulation micro-steps; 32\,768 tokens total.
    Standard deviations $\leq 0.13$\,s (CV~$<$~1.7\%).}
  \label{tab:model-abs-times}
  \footnotesize
  \begin{tabular}{l|ccc|ccc|ccc}
    \toprule
    \rowcolor{tableheader}
      & \multicolumn{3}{c|}{\textbf{Fused (s)}}
      & \multicolumn{3}{c|}{\textbf{Eager (s)}}
      & \multicolumn{3}{c}{\textbf{PEFT (s)}} \\
    \rowcolor{tableheader} \textbf{Model}
      & \textbf{RTX} & \textbf{H200} & \textbf{B200}
      & \textbf{RTX} & \textbf{H200} & \textbf{B200}
      & \textbf{RTX} & \textbf{H200} & \textbf{B200} \\
    \midrule
    Qwen2.5-VL-32B & OOM  & 30.9 & 23.3 & OOM  & 37.1 & 28.5
                   & OOM  & 53.6 & 40.4 \\
    Qwen3-VL-32B   & OOM  & 33.4 & 25.4 & OOM  & 39.5 & 30.6
                   & OOM  & 55.4 & 42.3 \\
    Qwen3.5-27B    & 39.5 & 25.4 & 19.4 & 48.1 & 30.6 & 23.9
                   & 59.5 & 39.8 & 30.5 \\
    Gemma3-27B     & 41.6 & 27.5 & 21.4 & 50.0 & 32.7 & 25.8
                   & 63.6 & 44.1 & 33.4 \\
    Mistral-Sm-24B & 32.5 & 20.9 & 15.9 & 39.7 & 25.2 & 19.6
                   & 53.9 & 39.1 & 29.8 \\
    Qwen3-VL-8B   & 12.9 &  9.1 &  7.1 & 15.9 & 11.0 &  8.8
                   & 18.8 & 13.7 & 10.5 \\
    \bottomrule
  \end{tabular}
\end{table}

\begin{figure}[!tbp]
  \centering
  \includegraphics[width=0.95\linewidth]{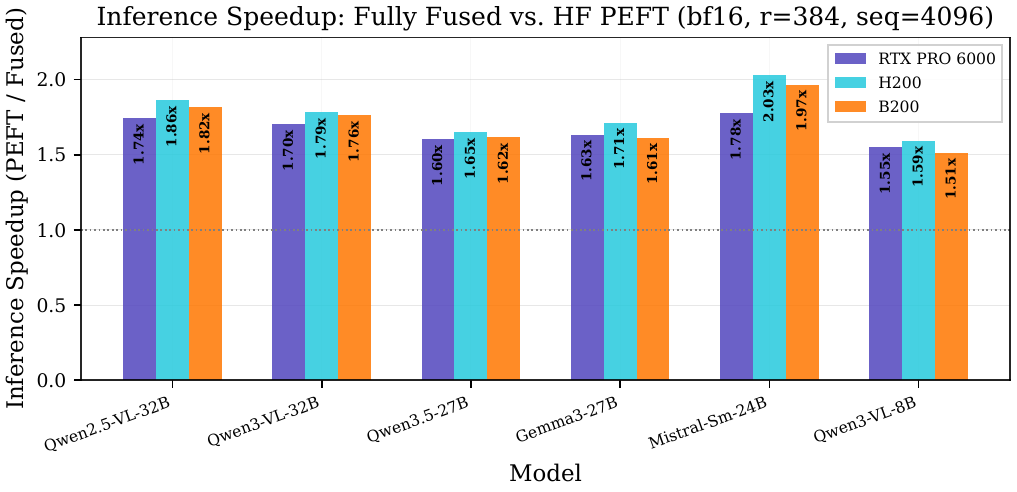}
  \caption{Inference speedup: $1.5$--$2.0$\tms{} over the HF~PEFT DoRA baseline.
    All six models run on all three GPUs, including 32B on RTX~6000~PRO
    (96\,GB) that OOM during gradient computation.}
  \label{fig:model-inference}
\end{figure}

\paragraph{Inference.}
Inference speedup is higher than gradient computation:
$1.5$--$2.0$\tms{} over PEFT, $1.14$--$1.20$\tms{} over eager
(Figure~\ref{fig:model-inference}), because the forward pass concentrates
the compose savings without dilution from backward-pass work.
RTX~6000~PRO runs inference on all six models including 32B
(84--88\,GB peak), which OOM during gradient computation.

\begin{figure}[!tbp]
  \centering
  \includegraphics[width=\linewidth]{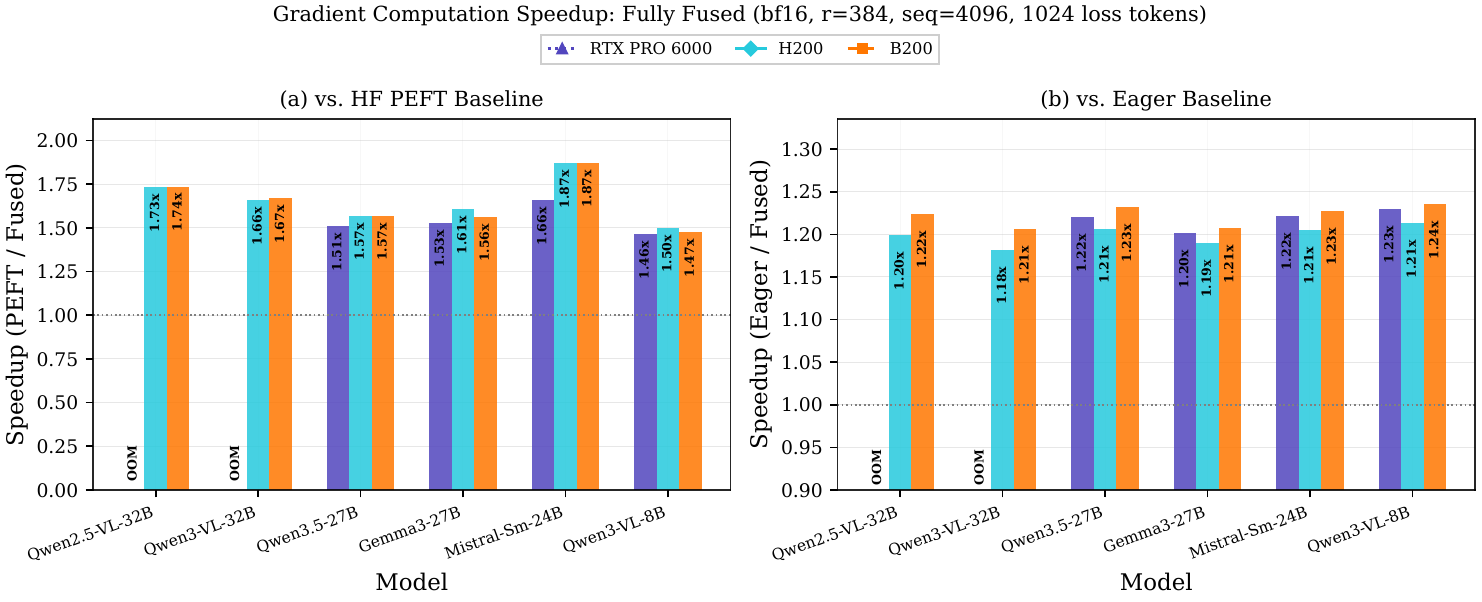}
  \caption{Gradient-computation speedup across six VLMs on three GPUs
    (bf16, $r\!=\!384$, seq=4096).  (a) Fused vs.\ the HF~PEFT DoRA baseline:
    $1.46$--$1.87$\tms{}.  (b) Fused vs.\ eager:
    $1.18$--$1.24$\tms{}.  32B models OOM on RTX~6000~PRO under all
    configurations.}
  \label{fig:model-training}
\end{figure}

\paragraph{High-rank scaling.}
Table~\ref{tab:high-rank} validates the high-rank framing at
$r\!=\!384$, $512$, and $768$.  Speedup vs.\ PEFT DoRA \emph{increases} with rank
for the 32B model ($1.66$\tms{} $\to$ $1.74$\tms{}) because PEFT's materialization
cost grows with $r$, while the factored norm's rank-dependent overhead
($\bmat{U}$ and $\bmat{G}$) remains small.  Speedup vs.\ eager decreases
modestly ($1.18$\tms{} $\to$ $1.14$\tms{}) as larger LoRA matmuls dilute the
compose kernel's contribution.

\begin{table}[!tbp]
  \centering
  \caption{Speedup vs.\ the HF~PEFT DoRA baseline grows with rank; speedup
    vs.\ eager decreases modestly (H200, bf16, seq=4096, 20 repeats).}
  \label{tab:high-rank}
  \small
  \begin{tabular}{llcc|cc}
    \toprule
    \rowcolor{tableheader} & & \multicolumn{2}{c|}{\textbf{vs.\ PEFT DoRA}}
      & \multicolumn{2}{c}{\textbf{vs.\ Eager}} \\
    \rowcolor{tableheader} \textbf{Model} & \textbf{Rank}
      & \textbf{Grad.} & \textbf{Infer.}
      & \textbf{Grad.} & \textbf{Infer.} \\
    \midrule
    \multirow{3}{*}{Qwen3.5-27B}
      & 384 & 1.57\tms{} & 1.65\tms{} & 1.21\tms{} & 1.16\tms{} \\
      & 512 & 1.61\tms{} & 1.68\tms{} & 1.18\tms{} & 1.14\tms{} \\
      & 768 & 1.53\tms{} & 1.59\tms{} & 1.15\tms{} & 1.11\tms{} \\
    \midrule
    \multirow{3}{*}{Qwen3-VL-32B}
      & 384 & 1.66\tms{} & 1.78\tms{} & 1.18\tms{} & 1.14\tms{} \\
      & 512 & 1.70\tms{} & 1.82\tms{} & 1.16\tms{} & 1.12\tms{} \\
      & 768 & 1.74\tms{} & 1.87\tms{} & 1.14\tms{} & 1.10\tms{} \\
    \bottomrule
  \end{tabular}
\end{table}

\FloatBarrier
\subsection{Why Dense (B@A) Is Not Enough}
\label{sec:dense-ba}

\begin{figure}[!tbp]
  \centering
  \includegraphics[width=\linewidth]{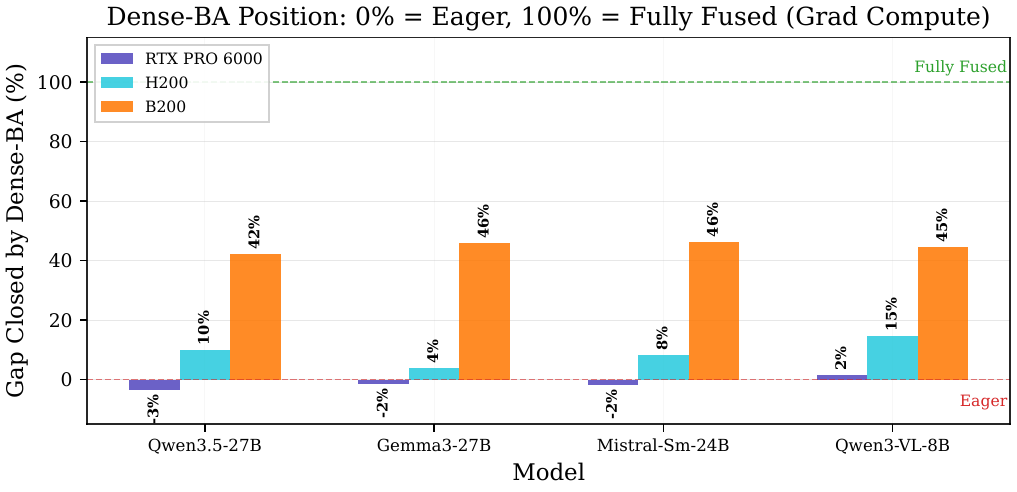}
  \caption{Dense~(B@A) position in the eager-to-fused gap (0\%~=~eager,
    100\%~=~fused).  Negative values: dense~(B@A) is \emph{slower} than eager.
    The benefit is GPU-bandwidth-sensitive; the factored approach is robust.}
  \label{fig:dense-ba}
\end{figure}

Computing \texttt{lora\_B.weight @ lora\_A.weight} directly (the most obvious
fix) eliminates the identity matrix but still materializes the full
$[d_\text{out}, d_\text{in}]$ product.  Figure~\ref{fig:dense-ba} shows that
dense~(B@A) captures 0\% of the eager-to-fused gap on some model/GPU
combinations and is sometimes \emph{slower} than the eager baseline.
Dense~(B@A) also uses 1--2\,GB more peak VRAM than fused on all tested models.
The full factored norm is necessary for consistent gains across GPU
architectures.

\FloatBarrier
\subsection{Compose Kernel Performance}

\begin{figure}[!tbp]
  \centering
  \includegraphics[width=\linewidth]{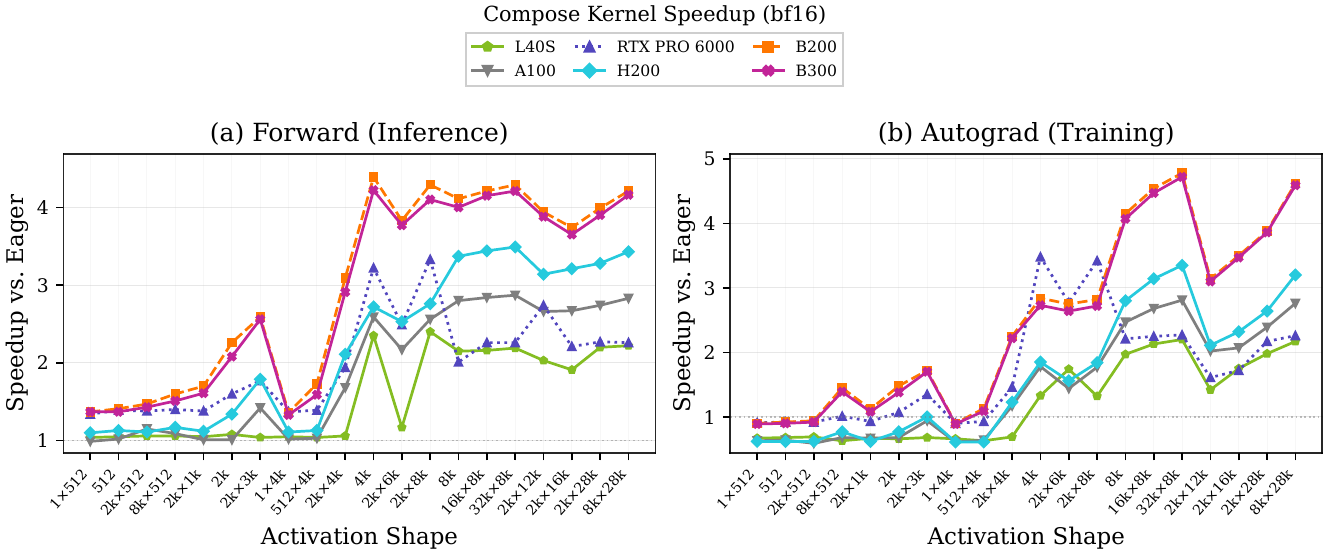}
  \caption{Compose kernel speedup vs.\ eager (bf16) across six GPUs.
    (a)~Forward: $1.5$--$4.5$\tms{}.  (b)~Autograd: gains compound
    with activation size.}
  \label{fig:compose}
\end{figure}

Figure~\ref{fig:compose} shows compose speedup across activation sizes on
six GPUs.  Geometric mean forward speedup (bf16, all 20 shapes):
$2.70$\tms{} B200, $2.62$\tms{} B300, $2.00$\tms{} H200,
$1.92$\tms{} RTX~6000~PRO, $1.73$\tms{} A100, $1.47$\tms{} L40S.
The consistency from GDDR6 (0.86\,TB/s) to HBM3e (7.7\,TB/s)
confirms the gains derive from reduced memory traffic rather than
architecture-specific effects.

\begin{figure}[!tbp]
  \centering
  \includegraphics[width=\linewidth]{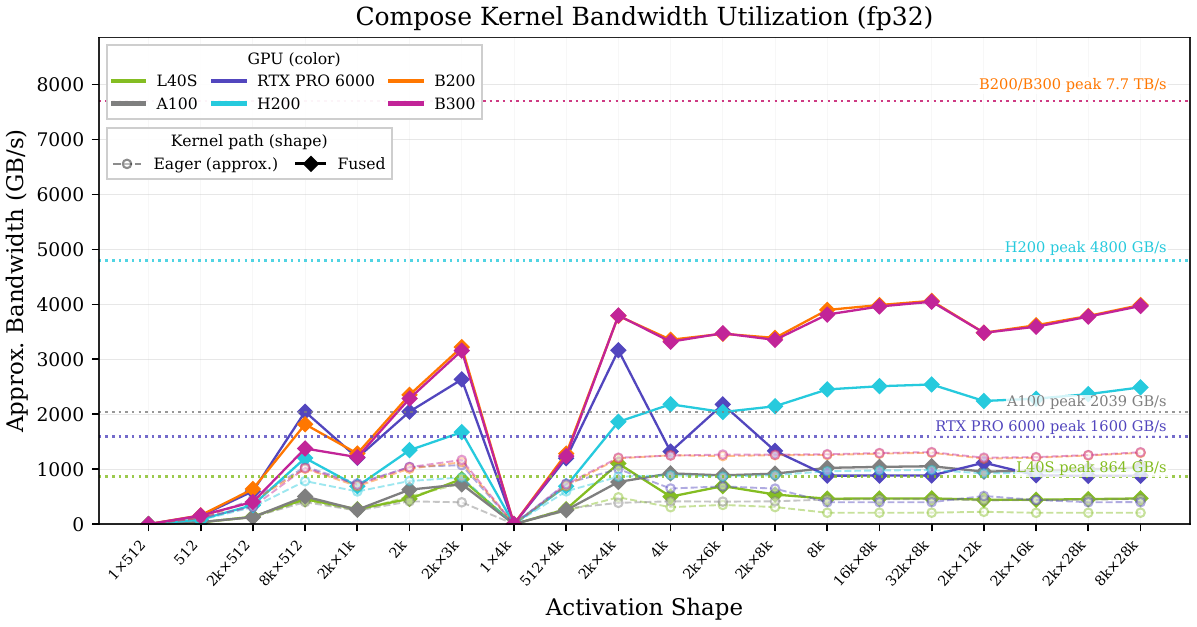}
  \caption{Bandwidth utilization (fp32, six GPUs).  Fused approaches
    ${\sim}50\%$ of peak on all architectures; eager values are approximate
    lower bounds.}
  \label{fig:bandwidth}
\end{figure}

\paragraph{Bandwidth utilization.}
The fused kernel achieves 3950--4070\,GB/s on B200/B300 (${\sim}53\%$ of peak),
2490--2540\,GB/s on H200 (${\sim}53\%$), 1040--1050\,GB/s on A100 (${\sim}52\%$),
880--890\,GB/s on RTX~6000~PRO (${\sim}55\%$), and 460--470\,GB/s on L40S
(${\sim}54\%$) at the largest shapes (Figure~\ref{fig:bandwidth}).
On B200, the eager path reaches only 17\% of peak, yielding the largest
absolute bandwidth gap.  Throughput scales nearly linearly with peak bandwidth
across the full 0.86--7.7\,TB/s range, confirming these kernels are
memory-bandwidth-bound.

\FloatBarrier
\subsection{Backward Kernel Performance}

\begin{figure}[!tbp]
  \centering
  \includegraphics[width=\linewidth]{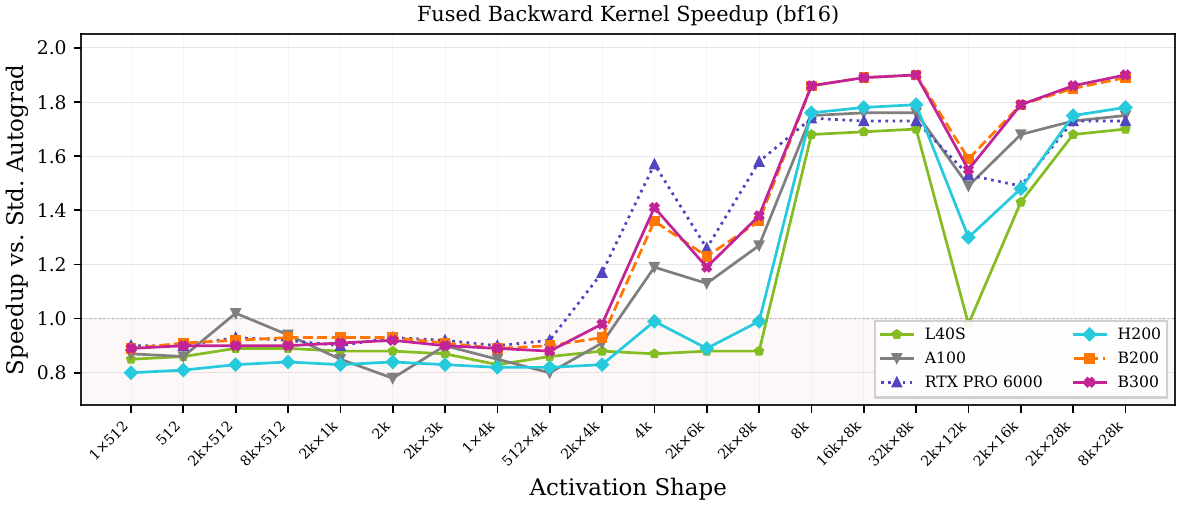}
  \caption{Backward speedup (bf16).  Below ${\sim}4096 \times 4096$, launch
    overhead dominates; above ${\sim}8192 \times 8192$, fused wins on all GPUs.}
  \label{fig:backward}
\end{figure}

The backward kernel shows a clear crossover: below ${\sim}2048 \times 6144$
(rows \tms{} $d_\text{out}$), launch overhead dominates and fused can trail
eager (0.88--0.99\tms{}); above ${\sim}8192 \times 8192$, fused wins on all
six GPUs (Figure~\ref{fig:backward}).  Geometric mean speedup (bf16, all
shapes): $1.23$\tms{} B200, $1.22$\tms{} B300/RTX~6000~PRO, $1.16$\tms{} A100,
$1.08$\tms{} H200, $1.06$\tms{} L40S.  Gradient correctness: fp32
$d_\text{lora}$ and $d_\text{base}$ match the eager baseline at tolerance
floor; $d_\text{mag}$ shows $\leq 2.14 \times 10^{-4}$ difference due to the
separate reduction path.

\FloatBarrier
\subsection{Norm Memory Reduction}

\begin{figure}[!tbp]
  \centering
  \includegraphics[width=\linewidth]{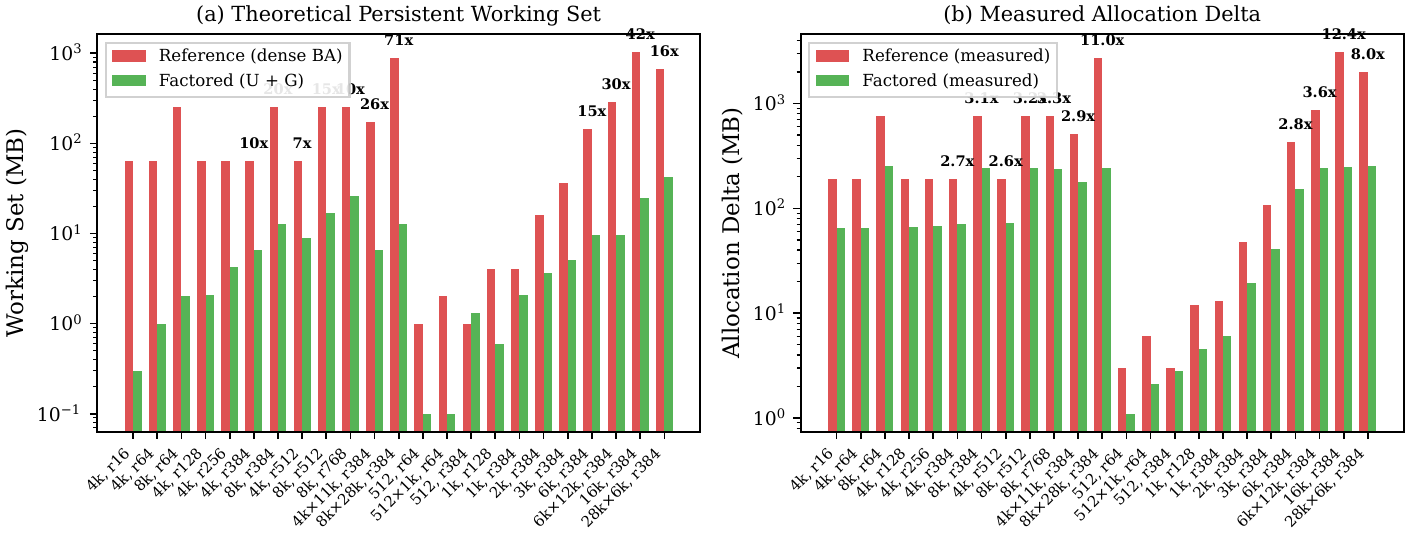}
  \caption{Norm memory reduction.  (a) Theoretical persistent working set.
    (b) Measured allocator delta.  The MoE shape $8192 \times 28672$ achieves
    $11$\tms{} measured reduction.}
  \label{fig:norm-memory}
\end{figure}

\begin{figure}[!tbp]
  \centering
  \includegraphics[width=\linewidth]{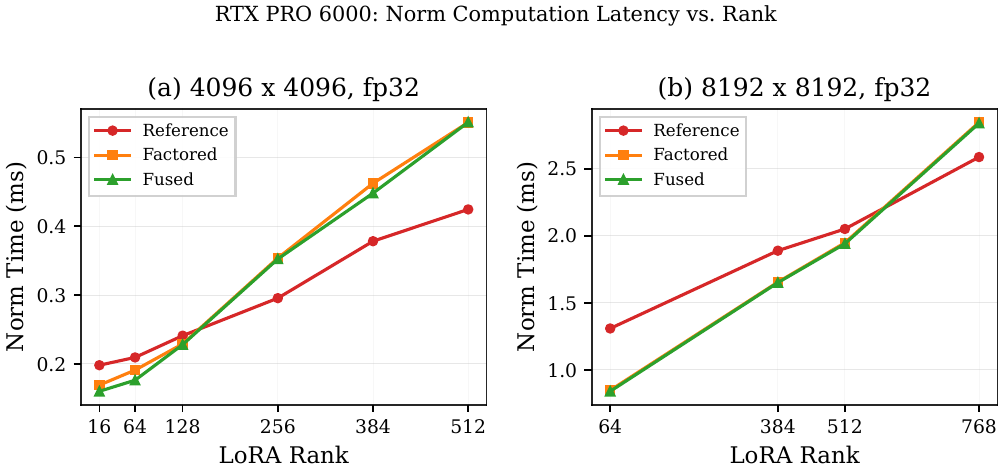}
  \caption{Norm latency vs.\ rank (RTX~6000~PRO, fp32).  The PEFT time is
    constant in $r$; factored scales linearly.  At $r \leq 128$, factored
    matches the reference due to reduced memory traffic.}
  \label{fig:norm-time}
\end{figure}

Figure~\ref{fig:norm-memory} and Table~\ref{tab:norm-memory} show both
theoretical and measured memory reductions.  The $8192 \times 28672$ MoE shape
achieves $11$\tms{} measured reduction.  The factored norm's latency tradeoff
(Figure~\ref{fig:norm-time}) is hardware-dependent: on RTX~6000~PRO, factored
matches or outperforms the reference at $r \leq 384$ for $8192 \times 8192$
matrices.

\begin{table}[!tbp]
  \centering
  \caption{Norm memory: measured allocation delta and theoretical reduction
    (fp32, H200).  Measured reductions are smaller than theoretical because
    they include the rank-independent base-norm transient
    (\S\ref{sec:complexity-detail}).}
  \label{tab:norm-memory}
  \small
  \begin{tabular}{llrrrc}
    \toprule
    \rowcolor{tableheader} \textbf{Shape} & \textbf{Rank}
      & \textbf{PEFT} & \textbf{Factored}
      & \textbf{Meas.\ \tms{}} & \textbf{Theory \tms{}} \\
    \midrule
    $4096 \times 4096$   & 64  & 192\,MB  & 65\,MB  & 3.0\tms{} & 63\tms{} \\
    $4096 \times 4096$   & 384 & 192\,MB  & 71\,MB  & 2.7\tms{} & 9.8\tms{} \\
    $4096 \times 4096$   & 512 & 192\,MB  & 73\,MB  & 2.6\tms{} & 7.1\tms{} \\
    $8192 \times 8192$   & 384 & 768\,MB  & 245\,MB & 3.1\tms{} & 20.4\tms{} \\
    $8192 \times 8192$   & 512 & 768\,MB  & 241\,MB & 3.2\tms{} & 15.1\tms{} \\
    $8192 \times 8192$   & 768 & 768\,MB  & 236\,MB & 3.2\tms{} & 9.8\tms{} \\
    $4096 \times 11008$  & 384 & 516\,MB  & 179\,MB & 2.9\tms{} & 26.2\tms{} \\
    $8192 \times 28672$  & 384 & 2688\,MB & 245\,MB & 11.0\tms{} & 71.3\tms{} \\
    \bottomrule
  \end{tabular}
\end{table}

\FloatBarrier
\subsection{Memory Profile}

\begin{figure}[!tbp]
  \centering
  \includegraphics[width=\linewidth]{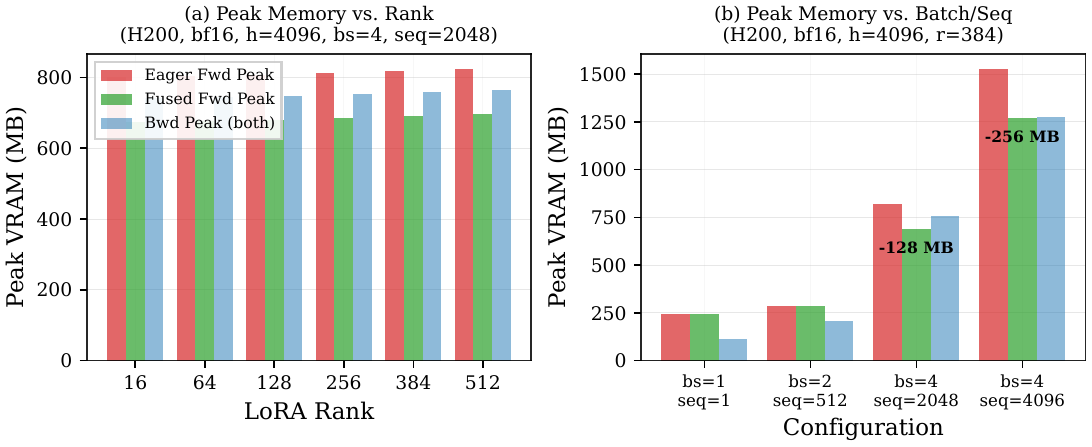}
  \caption{Memory profile (H200, bf16, $d\!=\!4096$, bs=4, seq=2048).
    (a) Fused reduces forward peak by 64\,MB.
    (b) Savings grow with batch\tms{}seq; backward peak is unchanged.}
  \label{fig:memory}
\end{figure}

The fused backward path reduces forward peak VRAM by eliminating intermediate
materialization while maintaining identical backward peak
(Figure~\ref{fig:memory}).  At the model level
(Table~\ref{tab:model-memory-comparison}), fused uses 0.1--1.0\,GB less peak
VRAM than eager and 1.2--6.7\,GB less than PEFT.
Dense~(B@A) uses more peak VRAM than fused on all models.

\begin{table}[!tbp]
  \centering
  \caption{Model-level peak VRAM (GB).  Fused uses less than all baselines
    on every model.  32B models OOM on RTX~6000~PRO.}
  \label{tab:model-memory-comparison}
  \small
  \begin{tabular}{llccc}
    \toprule
    \rowcolor{tableheader} \textbf{Model} & \textbf{Method}
      & \textbf{RTX} & \textbf{H200} & \textbf{B200} \\
    \midrule
    \multirow{4}{*}{Qwen2.5-VL-32B}
      & Eager & OOM & 99.2 & 99.2 \\
      & Fused & OOM & 98.4 & 98.4 \\
      & Dense (B@A) & OOM & 100.6 & 100.6 \\
      & PEFT & OOM & 103.5 & 103.5 \\
    \midrule
    \multirow{4}{*}{Mistral-Sm-24B}
      & Eager & 71.6 & 71.7 & 71.6 \\
      & Fused & 70.6 & 70.7 & 70.6 \\
      & Dense (B@A) & 73.3 & 73.4 & 73.3 \\
      & PEFT & 77.3 & 77.4 & 77.3 \\
    \midrule
    \multirow{4}{*}{Qwen3-VL-8B}
      & Eager & 29.8 & 29.9 & 29.8 \\
      & Fused & 29.5 & 29.5 & 29.5 \\
      & Dense (B@A) & 30.1 & 30.2 & 30.1 \\
      & PEFT & 30.7 & 30.8 & 30.7 \\
    \midrule
    \multicolumn{5}{l}{\footnotesize Full table (all 6 models) in
      Appendix~\ref{app:full-memory}.} \\
    \bottomrule
  \end{tabular}
\end{table}

\FloatBarrier
\subsection{Cross-Architecture Consistency}
\label{sec:model-perf}

Table~\ref{tab:summary} summarizes microbenchmark speedups across all six GPUs.
Model-level eager/fused speedups range from $1.18$\tms{} to $1.24$\tms{} with
cross-GPU CV~$<$~2\%, providing stronger statistical evidence than additional
repeats on a single GPU.

\begin{table}[!tbp]
  \centering
  \caption{Geometric mean microbenchmark speedups (all shapes, 200 repeats).
    Norm memory $0.8$\tms{} in bf16 means factored uses \emph{more} memory
    for the isolated norm due to fp32 accumulation transients
    (\S\ref{sec:complexity-detail}).}
  \label{tab:summary}
  \small
  \begin{tabular}{lcccc}
    \toprule
    \rowcolor{tableheader}
      & \textbf{Compose fwd} & \textbf{Backward}
      & \textbf{E2E} & \textbf{Norm mem} \\
    \midrule
    L40S bf16       & 1.47\tms{} & 1.06\tms{} & 1.05\tms{} & 0.8\tms{} \\
    A100 bf16       & 1.73\tms{} & 1.16\tms{} & 1.07\tms{} & 0.8\tms{} \\
    RTX 6000 PRO bf16 & 1.92\tms{} & 1.22\tms{} & 1.08\tms{} & 0.8\tms{} \\
    H200 bf16       & 2.00\tms{} & 1.08\tms{} & 1.06\tms{} & 0.8\tms{} \\
    B200 bf16       & 2.70\tms{} & 1.23\tms{} & 1.08\tms{} & 0.8\tms{} \\
    B300 bf16       & 2.62\tms{} & 1.22\tms{} & 1.08\tms{} & 0.8\tms{} \\
    \midrule
    \multicolumn{5}{l}{\footnotesize fp32 rows in
      Appendix~\ref{app:synthetic-e2e}.} \\
    \bottomrule
  \end{tabular}
\end{table}

\paragraph{Fidelity.}
Cosine similarity between fused and eager final logits exceeds $0.9999$ for all
six models on all three GPUs ($\cos \geq 0.999996$ on HBM-class GPUs).
An earlier code version showed reduced fidelity on Gemma-3-12B
($\cos = 0.991$--$0.999$); the root cause was fusing the magnitude division into
Triton, which allowed FMA contraction and approximate \texttt{sqrt} to perturb
rounding at large activation scales.  De-fusing the division (\S\ref{sec:dispatch}),
adding store-reload barriers, and replacing the \texttt{sqrt} with inline PTX
resolved the discrepancy, improving fidelity to $\cos > 0.9999$ across all GPUs.

\FloatBarrier
\subsection{Convergence Equivalence}
\label{sec:convergence-equivalence}

\begin{figure}[!tbp]
  \centering
  \includegraphics[width=\linewidth]{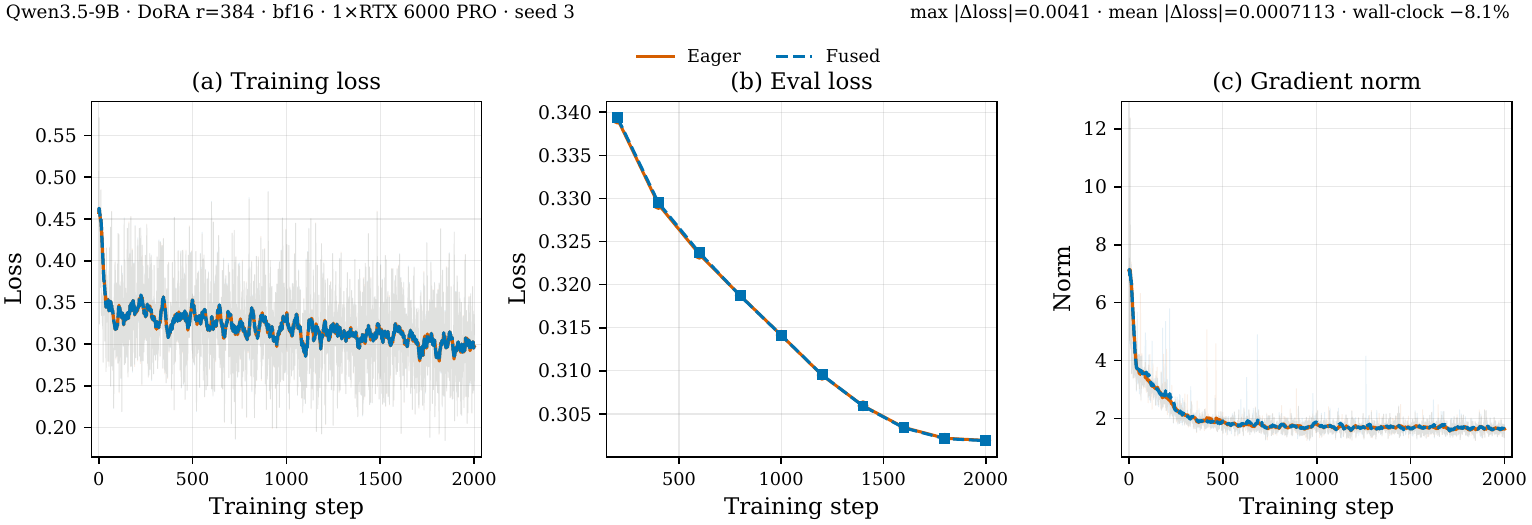}
  \caption{Convergence: eager vs.\ fused are visually indistinguishable
    (Qwen3.5-9B-Base, $r\!=\!384$, seed~3 of 3).
    (a) Training loss (25-step smoothing).  (b) Eval loss (200-step intervals).
    (c) Gradient norms.}
  \label{fig:convergence}
\end{figure}

To verify that fused kernels do not affect training dynamics, we trained
controlled SFT experiments on a length-filtered derivative of
MMFineReason-SFT-123K~\citep{lin2026mmfinereason} using
Qwen3.5-9B-Base, DoRA $r\!=\!384$, $\alpha\!=\!192$, rsLoRA, bf16, AdamW,
ZeRO-2, gradient checkpointing, $\text{bs}\!=\!3$, $\text{ga}\!=\!2$,
$\text{seq}\!=\!5120$, 2000 steps on a single RTX~6000~PRO,
using the SWIFT framework~\citep{zhao2024swift}, with three seeds
(\tms{} eager/fused $= 6$ runs).
Table~\ref{tab:convergence-multi} and Figure~\ref{fig:convergence} summarize
the results.

\begin{table}[!tbp]
  \centering
  \small
  \caption{Multi-seed convergence: eager vs.\ fused training loss
    (Qwen3.5-9B-Base, $r\!=\!384$, 2000 steps).  Grand mean per-step delta
    $7.1 \times 10^{-4}$; final eval losses agree to
    $< 1.5 \times 10^{-4}$.}
  \label{tab:convergence-multi}
  \begin{tabular}{@{}lccccc@{}}
    \toprule
    \rowcolor{tableheader} Seed & Steps & Mean $|\Delta|$
      & Max $|\Delta|$ & Eval $|\Delta|$
      & Wall (fused/eager) \\
    \midrule
    1 & 2000 & $7.2\!\times\!10^{-4}$ & $1.1\!\times\!10^{-2}$
      & $9.2\!\times\!10^{-5}$ & 330/362\,min \\
    2 & 2000 & $6.9\!\times\!10^{-4}$ & $3.3\!\times\!10^{-3}$
      & $1.4\!\times\!10^{-4}$ & 330/359\,min \\
    3 & 2000 & $7.1\!\times\!10^{-4}$ & $4.1\!\times\!10^{-3}$
      & $3.3\!\times\!10^{-5}$ & 330/359\,min \\
    \midrule
    \multicolumn{2}{@{}l}{\emph{Grand mean}}
      & $7.1\!\times\!10^{-4}$ & ---
      & $8.9\!\times\!10^{-5}$ & 330/360\,min \\
    \bottomrule
  \end{tabular}
\end{table}

The worst-case single-step delta ($1.1 \times 10^{-2}$, seed~1, step~398) is
a transient early-training divergence that does not propagate: by step~1000,
all deltas fall below $3.3 \times 10^{-3}$.  Gradient norms track identically,
confirming that the $d_\text{mag}$ reduction-ordering difference does not
accumulate over 2000 steps.

\paragraph{Wall-clock.}
The fused path completed 2000 steps in 330\,min compared with 360\,min for the eager baseline (8.3\%
reduction), consistent with the 21\% gradient-computation speedup diluted by
optimizer steps, data loading, and framework overhead.

\paragraph{Cross-model and cross-optimizer check.}
An additional pair on Qwen3-VL-8B-Instruct with Muon+AdamW ($r\!=\!256$,
single seed) showed consistent results: mean
$|\Delta\text{loss}| = 7.7 \times 10^{-4}$, final eval
$|\Delta| = 3.9 \times 10^{-5}$, 8.2\% wall-clock reduction.

\FloatBarrier
\section{Discussion}
\label{sec:discussion}

\subsection{Deployment Context}

The factored norm is particularly valuable when training and inference compete
for GPU memory.  Our GRPO~\citep{shao2024deepseekmath} pipeline co-locates
vLLM~\citep{kwon2023vllm} (tensor-parallel inference) alongside DoRA
fine-tuning ($r\!=\!384$) of a 38B VLM on 4\tms{}B200 (192\,GB each),
with large global batches under ZeRO-2 and gradient checkpointing.  After
vLLM reserves its KV-cache allocation, training headroom per GPU is tight;
the memory challenge is cumulative rather than catastrophic.  Each of the
500+ adapted modules re-materializes its norm temporaries during gradient
checkpointing recomputation, and the resulting transient allocations fragment
the caching allocator.  Cross-device bandwidth, already under pressure from
gradient all-reduce and tensor-parallel inference communication, leaves
little margin for the additional memory traffic of dense per-module
materialization.  The factored norm eliminates these transients, and we
observed no numerical drift attributable to fusion.  (This is an illustrative
anecdote and was not benchmarked under the methodology of \S\ref{sec:experiments}.)

\subsection{Tradeoffs and Limitations}

Table~\ref{tab:decision-guide} consolidates practitioner recommendations.

\begin{table}[!tbp]
  \centering
  \caption{Recommended configuration by scenario.}
  \label{tab:decision-guide}
  \small
  \begin{tabularx}{\linewidth}{@{}>{\raggedright\arraybackslash}p{3.2cm}l
    >{\raggedright\arraybackslash}X@{}}
    \toprule
    \rowcolor{tableheader} \textbf{Scenario} & \textbf{Config}
      & \textbf{Rationale} \\
    \midrule
    Training, CUDA, above crossover & Tier~1 (fused bwd)
      & Full speedup; 0.1--1.0\,GB \emph{less} VRAM than eager \\[3pt]
    Training, memory-constrained & Tier~1 + frozen mag
      & \texttt{inner} skipped; lower peak VRAM \\[3pt]
    Inference, CUDA & Tier~2 (fused fwd)
      & Compose speedup, no backward memory \\[3pt]
    CPU / no Triton & Tier~3 (eager)
      & Automatic fallback \\[3pt]
    $r \geq 384$, any single GPU & Factored norm
      & PEFT path 46--87\% slower \\[3pt]
    $r \leq 64$, $d \leq 4096$ & Either norm
      & Factored overhead minimal \\[3pt]
    Colocated train + infer & Factored norm
      & Dense temporaries compete with inference budget \\
    \bottomrule
  \end{tabularx}
\end{table}

\paragraph{Where fusion offers no advantage.}
Below ${\sim}2048 \times 6144$ activations, launch latency dominates;
the dispatch encodes this crossover conservatively.  On non-CUDA platforms,
Triton kernels are unavailable.

\paragraph{Fused backward VRAM.}
The fused backward saves one activation-sized tensor (\texttt{inner}) per
module, but the dual-output kernel also eliminates the forward-pass spike from
sequential ops.  Net effect: fused uses 0.1--1.0\,GB \emph{less} peak VRAM
than eager at the model level.  With frozen magnitude,
\texttt{inner} is skipped entirely.

\paragraph{Numerical precision.}
All PyTorch compose paths are bitwise identical.  Triton preserves the same
algebra but not bitwise equality (\S\ref{sec:dispatch}).  Residual drift
concentrates in $d_\text{mag}$ reductions rather than pointwise compose.
Convergence studies (\S\ref{sec:convergence-equivalence}) confirm these
differences do not accumulate.

\paragraph{Distributed training.}
DeepSpeed ZeRO-2/3 and FSDP1 are supported.  FSDP2/DTensor is not: the
factored norm assumes access to the full base weight $\bmat{W}$.  Extending to
FSDP2 would require distributed accumulation of the chunk-wise partial sums
followed by an all-reduce over the shard dimension; the per-row output
($[d_\text{out}]$) is small enough to replicate.  We leave this for future work.

\paragraph{Embedding formula correction.}
PEFT's embedding path computes only $g \odot \text{lora} \cdot s$, omitting
$(g{-}1) \odot \text{base}$.  Our implementation applies the full DoRA formula
consistently across all layer types.  No headline benchmarks include adapted
embeddings; checkpoints fine-tuned with PEFT's embedding path may require
re-fine-tuning or a legacy composition fallback.

\paragraph{Ablation.}
Model-level speedups reflect both contributions (factored norm + fused
kernels) jointly.  Microbenchmarks (Tables~\ref{tab:summary}
and~\ref{tab:norm-memory}) provide component-level measurements, and the
model-level eager-vs.-fused comparison provides a partial ablation of the
kernel-fusion contribution.  A fuller factorial ablation across additional
model families would strengthen the evidence.

\section{Related Work}
\label{sec:related}

\paragraph{Parameter-efficient fine-tuning.}
LoRA~\citep{hu2022lora} introduced low-rank adapter decomposition;
DoRA~\citep{liu2024dora} adds magnitude-direction separation.
rsLoRA~\citep{kalajdzievski2023rslora} provides rank-stabilized
scaling that interacts with our factored norm ($s$ appears in all
three terms of Equation~\ref{eq:factored-norm}).

\paragraph{DoRA variants.}
EDoRA~\citep{nasiri2025edora} reduces static parameter count via SVD;
DoRAN~\citep{truong2025doran} injects noise into the normalization
denominator.  Both address statistical efficiency rather than transient
memory; our optimization is complementary.
Chronicals~\citep{nair2026chronicals} and
LoRAFusion~\citep{zhu2026lorafusion} fuse LoRA-related operations but do not
target the DoRA-specific norm or composition.

\paragraph{Framework implementations.}
Every major framework we checked (HF~PEFT, torchtune, Unsloth, SWIFT,
LLaMA-Factory, Axolotl) uses the same \texttt{torch.eye} materialization
pattern.  Unsloth explicitly disables its custom kernels when DoRA is active;
orchestration frameworks delegate entirely to PEFT.  As of February~2026, no
existing framework avoids materializing the dense $\bmat{B}\bmat{A}$ product
(Appendix~\ref{app:framework-survey}).

\paragraph{Kernel fusion.}
FlashAttention~\citep{dao2022flashattention,dao2024flashattention2}
demonstrated that tiled, fused kernels improve both speed and memory for
attention.  Liger Kernel~\citep{hsu2024liger} applies similar principles to
cross-entropy, SwiGLU, and RMSNorm.  Our work targets the DoRA composition, a
simpler (element-wise with broadcasting) but equally memory-bound pattern.
The algebraic identity underlying the factored norm (expanding a sum-of-squares
into base, cross, and Gram terms) is standard in numerical linear algebra; our
contribution is its application to the DoRA-specific computation with dtype
discipline, chunking, and integration into the fused pipeline.

\paragraph{LLM-guided optimization.}
Meta's KernelAgent~\citep{kernelagent2025} confirmed our compose kernel is
near-roofline (89\% memory bandwidth SOL, 1.5\% improvement).  For the
backward, KernelAgent discovered a two-stage partial-reduction strategy that
fuses the $d_\text{mag}$ reduction, achieving $3.58$\tms{} over eager
(88.5\%~SOL) vs.\ our $1.06$--$1.23$\tms{}.  Our release prioritizes
drop-in compatibility and end-to-end wins across real models; integrating
that pattern is a direct avenue for future work.  KernelAgent's generated
listings are included in \texttt{code/kernelagent\_sols}.

\section{Conclusion}

We presented a systems implementation of DoRA: a factored norm that reduces
working memory from $\bigO(d_\text{out} \times d_\text{in})$ to
$\bigO(d_\text{out} \times r + r^2)$, and fused Triton kernels that collapse
multi-step composition into single-pass GPU operations.

On six 8--32B VLMs, the fused implementation is $1.5$--$2.0$\tms{} faster than
HF~PEFT's DoRA implementation for inference, and $1.5$--$1.9$\tms{} faster for
gradient computation (optimizer step excluded), with up to $7$\,GB lower peak
VRAM.  Microbenchmarks on six GPUs spanning four
architecture generations confirm $1.5$--$2.7$\tms{} compose-kernel speedup.
Fidelity holds at three levels: operator tests within quantization-aware bounds,
final-logit $\cos > 0.9999$, and matched training curves across seeds.

\paragraph{Known limitations.}
FSDP2 is unsupported.  Convergence validation covers two model families, two
optimizers, and one dataset in the SFT regime; generalization to RL pipelines
remains to be confirmed.  Model-level benchmarks cover three of six GPUs;
L40S, A100, and B300 have microbenchmark coverage only.  The dispatch crossover
is an empirical heuristic that may need retuning for future hardware.

\section*{Data Availability}

All source code, benchmark scripts, raw JSON results, Triton autotune caches,
and figure generation scripts are available at
\url{https://github.com/sockeye44/dorafactors} (tag \texttt{v1.0}).
The convergence validation uses a public dataset
(MMFineReason-SFT-123K;~\citealt{lin2026mmfinereason}) for fully reproducible
confirmation.  The authors declare no competing interests.

\section*{Acknowledgements}

This work was developed through extensive collaborative programming with
Claude Opus~4.6 (Anthropic), which contributed to kernel implementation,
test design, numerical analysis, and iterative debugging.  The authors take
full responsibility for the accuracy and integrity of the work.

\bibliography{references}

\clearpage
\appendix

\section{Forward Contract and Execution Semantics}
\label{app:forward-contract}

\begin{papercallout}{Forward Contract and Execution Semantics}
\noindent\textbf{1. Module Interface \& Compose Semantics}
\begin{itemize}[noitemsep, topsep=2pt]
  \item \textbf{Output:} The module computes a delta
    $\Delta \bmat{Y}$; the caller applies
    $\bmat{Y} = \bmat{Y}_\text{base} + \Delta \bmat{Y}$.
  \item \textbf{Compose Equation:}
    $\Delta \bmat{Y} = g \odot (s\bmat{X}\bmat{A}^\top\bmat{B}^\top)
    + (g - 1) \odot \bmat{Y}_\text{base}$.
\end{itemize}

\medskip
\noindent\textbf{2. Norm Policy}
\begin{itemize}[noitemsep, topsep=2pt]
  \item Recomputed every forward pass; never cached across steps.
  \item Detached (no gradient flow), per~\citet{liu2024dora} \S4.3.
  \item Accumulated in FP32 with autocast disabled.
  \item $\epsilon = 10^{-12}$ (fp32/fp64) or $10^{-6}$ (bf16/fp16).
  \item Bias subtracted before compose, re-added after.
\end{itemize}
\par\medskip
\textit{Formal contract for clean-room replication.}
\end{papercallout}

\section{Implementation Details}
\label{app:impl}

\paragraph{Chunk alignment.}
The chunk size aligns to 64 elements on CUDA/XPU devices for Tensor Core
MMA alignment on all NVIDIA architectures since Volta.

\paragraph{Environment variables.}
\nolinkurl{PEFT_DORA_FUSED} (0 = force eager),
\nolinkurl{PEFT_DORA_FUSED_BACKWARD} (1 = force fused bwd, 0 = disable,
unset = auto),
\nolinkurl{PEFT_DORA_NORM_CHUNK_MB} and
\nolinkurl{PEFT_DORA_FWD_CHUNK_MB} (override 256\,MB defaults).

\paragraph{Scale-is-zero fast path.}
When $s = 0$, cross and ba\_sq are skipped; $\bmat{U}$ and $\bmat{G}$ are
not allocated.

\paragraph{Dtype-aware epsilon.}
$10^{-12}$ for fp32/fp64; $10^{-6}$ for bf16/fp16.  For fp16 (max
$\approx 65504$), $\varepsilon = 10^{-6}$ limits the quotient to ${\sim}10^6$,
reducing saturation risk.

\paragraph{Compose kernel autotuning.}
\label{app:autotune}
RPP=1 is selected in 95\% of autotuned entries (1149/1206).  Exact config
agreement between GPUs is ${\sim}9\%$, confirming per-device autotuning is
essential.

\paragraph{Chunked dropout path.}
When dropout is active, \texttt{\_compose\_with\_base\_chunks} iterates
over output-dimension slices with adaptive sizing, decorated with
\texttt{@dynamo\_disable} to avoid runaway recompilations.

\paragraph{Magnitude broadcast shape guard.}
A shape guard gates Triton kernel dispatch on whether the magnitude vector
broadcasts exclusively along the last dimension of the activation tensor.
The Triton compose kernel treats magnitude as a 1-D vector along the last
dimension; Conv-style shapes like $[1, C, 1, 1]$ applied to $[N, C, H, W]$
activations would violate this assumption.  The guard checks both element
count and last-dimension alignment; failing shapes route to the PyTorch
fallback.

\paragraph{Custom op for torch.compile.}
The registered backward uses PyTorch (not Triton) because AOTAutograd
traces with FakeTensors.  Eager training uses Triton for both forward
and backward; compiled training uses Inductor to fuse the PyTorch
backward graph.

\section{Kernel Specifications}
\label{app:kernel-spec}

This appendix provides exact specifications for the three Triton kernels
and the PyTorch magnitude division stage, including casting points,
fused operations, shape constraints, and reduction ordering,
to support a clean-room reimplementation.

\paragraph{1. Compose Forward kernel.}
Fuses $(g-1) \odot \text{base} + g \odot s \odot \text{lora}$ in one pass.
Inputs: base $[\text{bs}, \text{seq}, d_\text{out}]$,
lora $[\text{bs}, \text{seq}, d_\text{out}]$,
$g$ $[d_\text{out}]$, $s$ (scalar).
Output: delta $[\text{bs}, \text{seq}, d_\text{out}]$.
All tensors in input dtype (fp16/bf16/fp32); no intermediate dtype cast.
$g$ is broadcast along all but the last dimension.

\paragraph{2. Compose Backward kernel.}
Fuses $d_\text{lora} = g \cdot s \cdot d_\text{out}$ and
$d_\text{base} = (g-1) \cdot d_\text{out}$ in a single Triton pass.
$d_\text{mag}$ is computed separately via a \texttt{.sum()} reduction over
the batch/sequence dimensions on the \texttt{inner} activation; this avoids
non-deterministic \texttt{tl.atomic\_add} ordering.

\paragraph{3. Norm Assembly kernel (norm-only).}
Inputs: base\_sq $[d_\text{out}]$, cross $[d_\text{out}]$, ba\_sq $[d_\text{out}]$
(all fp32), \texttt{two\_s} (scalar, $= 2s$, precomputed in fp64),
\texttt{s2} (scalar, $= s^2$, precomputed in fp64).
Computes $w_\text{norm} = \sqrt{\max(\text{base\_sq} + \texttt{two\_s} \cdot
\text{cross} + \texttt{s2} \cdot \text{ba\_sq},\, 0)}$ in fp32 with
store-reload barriers after each multiply-add to prevent FMA fusion, exactly
reproducing PyTorch's separate-kernel evaluation order.  The clamp preserves
NaN semantics (matching \texttt{torch.clamp\_min}, which propagates NaNs per
IEEE~754) rather than collapsing NaNs to zero.
The square root uses
inline PTX \texttt{sqrt.rn.f32} for IEEE~754 correctly-rounded results
(Triton's \texttt{tl.sqrt} compiles to \texttt{sqrt.approx.ftz.f32} on SM90).
The kernel returns the result in the input dtype.  In default mode, it uses a
fixed block size of 256 (norm kernels are launch-latency bound; see
Appendix~\ref{app:autotune}); comprehensive autotuning over 36 configurations
(block sizes 32--2048) is available for new GPU architectures.  If future
Triton versions change the lowering of \texttt{tl.sqrt} to IEEE-compliant
rounding, the inline PTX can be removed; the Tier-3 eager fallback provides
a portable alternative on any platform.

\paragraph{4. Magnitude division (PyTorch).}
The division $g = \bmat{m} / \max(w_\text{norm}, \varepsilon)$ is always
computed in PyTorch after the norm assembly kernel returns.  This ensures
identical precision regardless of whether the Triton or PyTorch norm path
was used, at the cost of one additional element-wise kernel launch
(negligible relative to surrounding matmuls).

\paragraph{Shape constraints.}
$d_\text{out}$ must be divisible by \texttt{BLOCK\_SIZE} (128).  The magnitude
vector must broadcast only along the last dimension of the activation; other
broadcast shapes (e.g., $[1, C, 1, 1]$ applied to $[N, C, H, W]$) route to the
Tier-3 eager fallback.  Non-contiguous input tensors also fall back to
Tier~3.

\paragraph{Tested compatibility matrix.}
Table~\ref{tab:compat} summarizes the integration points explicitly
tested, with notes on scope and caveats.  ``Tested'' indicates the feature was
exercised in benchmarks or convergence runs reported in this paper;
``CI only'' indicates coverage via the test suite (1041 tests) but not in
model-level experiments.

\begin{table}[!htbp]
  \centering
  \caption{Compatibility matrix.  Scope: \emph{Bench} = model-level benchmarks,
    \emph{Conv} = convergence runs, \emph{CI} = operator-level test suite.}
  \label{tab:compat}
  \footnotesize
  \begin{tabular}{lcc>{\raggedright\arraybackslash}p{4.8cm}}
    \toprule
    \rowcolor{tableheader} \textbf{Feature} & \textbf{Status} & \textbf{Scope} & \textbf{Notes} \\
    \midrule
    Mixed-precision bf16       & \cmark & Bench+Conv & All model benchmarks and convergence runs \\
    Mixed-precision fp16       & \cmark & CI         & Operator tests only; no model-level fp16 runs \\
    Gradient checkpointing     & \cmark & Bench+Conv & All model benchmarks use gradient checkpointing \\
    ZeRO-2                     & \cmark & Conv       & All convergence runs use ZeRO-2 \\
    ZeRO-3                     & \cmark & CI         & Parameter gathering tested; no model-level runs \\
    FSDP1                      & \cmark & CI         & \texttt{summon\_full\_params} path tested \\
    FSDP2 / DTensor            & \xmark & ---        & Not supported; see \S\ref{sec:discussion} \\
    \texttt{torch.compile}     & \cmark & CI         & Graph-break-free when $p\!=\!0$; see \S\ref{sec:discussion} \\
    Linear layers              & \cmark & Bench+Conv & Primary target; all benchmarks \\
    Conv1d / Conv2d / Conv3d   & \cmark & CI         & Correctness tests; no perf benchmarks \\
    Embedding layers           & \cmark & CI         & Formula correction (\S\ref{sec:discussion}); no perf benchmarks \\
    \bottomrule
  \end{tabular}
\end{table}

\section{Reproducibility}
\label{app:repro}

\paragraph{Code and data.}
All source code, benchmark scripts, raw JSON results, Triton autotune caches,
and figure generation scripts are available at
\url{https://github.com/sockeye44/dorafactors} (tag \texttt{v1.0}).
The patched PEFT module is included as a git submodule
(\texttt{vendor/dorafactors-peft}, branch \texttt{v1}); cloning with
\texttt{-{}-recurse-submodules} fetches it automatically.
Alternatively, the patch can be reconstructed via \texttt{git apply hf.patch}
against upstream PEFT commit
\texttt{20a9829}\footnote{PEFT commit: \href{https://github.com/huggingface/peft/commit/20a9829}{20a9829} (\texttt{v0.18.0.rc0}, 2025-09-16).}.
All commands below assume the repository root as working directory.

\paragraph{Software environment.}
All benchmarks were run under a single, pinned software stack:
PyTorch 2.10.0+cu130 (built against CUDA~13.0 for compatibility),
Triton 3.6.0, Transformers 5.2.0,
CUDA toolkit 13.1 (ptxas V13.1.115), driver 580.126.09,
Python 3.12.12 on Linux 6.8.0 (Ubuntu 22.04, glibc 2.35).
The exact environment is published as a Docker image\footnote{Docker image:
\url{https://hub.docker.com/r/alexazel/dorafactors-env}. Tag:
\texttt{cu131-pt210-vllm-t52-base}.} for full-stack reproducibility;
a \texttt{code/requirements.txt} is also included.

\paragraph{Memory measurement methodology.}
We report three complementary memory metrics, each appropriate to a different
level of analysis:
\begin{itemize}
  \item \textbf{Allocator peak}
    (\texttt{torch.cuda.max\_memory\_allocated()}): the maximum bytes
    actually allocated by PyTorch's caching allocator.  Used for
    microbenchmark memory deltas (Tables~\ref{tab:complexity} and
    \ref{tab:norm-memory}), measured after
    \texttt{reset\_peak\_memory\_stats()} and \texttt{empty\_cache()} to
    isolate a single operation's footprint.
  \item \textbf{Working-set delta}
    (\texttt{max\_memory\_allocated} $-$ \texttt{baseline\_allocated}):
    the peak minus the model's quiescent allocation, capturing the true
    transient overhead of DoRA's forward/backward pass.  Used for
    model-level gradient-computation analysis (\S\ref{sec:dense-ba},
    Table~\ref{tab:model-perf-comparison}).
  \item \textbf{Reserved VRAM} (\texttt{memory\_reserved}): the amount
    of memory the GPU physically withholds from other processes, including caching
    allocator fragmentation overhead.  Used for peak VRAM
    comparison (Table~\ref{tab:model-memory-comparison}) because it
    determines whether colocated workloads can share the device.
\end{itemize}
\noindent Every memory claim in this paper specifies both the metric and the
dtype (fp32 vs.\ bf16) to avoid conflation.

\par\noindent\textbf{Microbenchmark reproduction.}
\begin{lstlisting}
# 200 repeats, extended shapes, bf16
python code/bench_dora_comprehensive.py \
  --shapes extended --repeats 200 --warmup 10 \
  --dtype bf16 --json-out results.json
\end{lstlisting}
Each run produces a self-contained JSON file with per-test timing
distributions (200 samples), memory measurements, and pre-computed summary
statistics.  The \texttt{--shapes extended} flag generates the 20 unique
activation shapes (60 entries across 3 ranks) used throughout this paper.

\par\noindent\textbf{Model identifiers.}
All model-level benchmarks use the following Hugging Face model IDs
(weights downloaded March~2026; exact file hashes in the JSON artifacts):
\begin{itemize}[noitemsep, topsep=2pt, leftmargin=1.5em]
  \item \texttt{Qwen/Qwen2.5-VL-32B-Instruct}
  \item \texttt{Qwen/Qwen3-VL-32B-Instruct}
  \item \texttt{Qwen/Qwen3.5-27B}
  \item \texttt{google/gemma-3-27b-it}
  \item \texttt{unsloth/Mistral-Small-3.2-24B-Instruct-2506}
  \item \texttt{Qwen/Qwen3-VL-8B-Instruct}
\end{itemize}

\par\noindent\textbf{Convergence validation dataset.}
\begin{sloppypar}%
\emergencystretch=2em
The convergence validation (\S\ref{sec:convergence-equivalence}) uses a
token-length-filtered subset of
OpenDataArena/\allowbreak{}MMFine\-Reason-SFT-123K-Qwen3-VL-235B-Thinking~\citep{lin2026mmfinereason},
repacked with mechanical field renames
(\texttt{question}$\to$\texttt{query},
\texttt{qwen3vl\_235b\_thinking\_response}$\to$\texttt{response})
and filtered to $\text{tok\_len} \leq 4096$.
The repacked dataset is published at
\nolinkurl{eyes-ml/\allowbreak{}MMFineReason-SFT-123K-\allowbreak{}Qwen3-VL-235B-Thinking-QR-max4096}
on Hugging~Face Hub; the filtering script is included in the repository
(\texttt{code/\allowbreak{}scripts/\allowbreak{}repack\_mmfinereason\_qr.py}).
\end{sloppypar}

\par\noindent\textbf{Convergence validation environment.}
Training uses
\href{https://github.com/modelscope/ms-swift}{SWIFT}~\citep{zhao2024swift} (commit \texttt{a807cb9}) with
PyTorch~2.10.0+cu130, Transformers~5.2.0, Triton~3.6.0,
DeepSpeed~0.18.6, Flash-Attention~2.8.3.
The full environment (including \texttt{qwen-vl-utils},
\texttt{mamba\_ssm}, \texttt{flash-linear-\allowbreak{}attention}) uses
the same Docker image as the benchmarks (see Software environment above)
with the additional training dependencies installed.

\par\noindent\textbf{Model benchmark reproduction.}
\begin{lstlisting}
# 6 models, r=384, loss_tokens=1024
python code/bench_dora_comprehensive.py \
  --suite models --rank 384 --batch 1 --seqlen 4096 \
  --grad-accum 8 --loss-tokens 1024 --repeats 20 \
  --json-out models.json
\end{lstlisting}

\paragraph{Figure regeneration.}
All figures can be regenerated from the included JSON artifacts:
\begin{lstlisting}
python paper/generate_figures.py
\end{lstlisting}
This produces 13 PDF figures in \path{paper/figures/} sourced from the
\path{code/bench_it6/} data directory (6~GPUs \tms{} 3~dtypes for
microbenchmarks, 3~GPUs for model-level).  The convergence figure
(Figure~\ref{fig:convergence}) is generated separately from TensorBoard
logs via \texttt{python} \path{paper/generate_training_figure.py}.

\paragraph{Test suite.}
The full test suite (1041 tests) has been validated on SM\,80 through
SM\,120 (Ampere--Blackwell); Triton kernel tests require SM\,$\geq$\,80:
\begin{lstlisting}
cp code/scripts/dora.reference_hf_peft.py \
   vendor/dorafactors-peft/docs/
cd vendor/dorafactors-peft
pytest tests/test_lora_variants.py \
       tests/tuners/lora/test_dora_fused.py \
       tests/tuners/lora/test_dora_math.py -v
\end{lstlisting}

\section{Full Model-Level Memory Table}
\label{app:full-memory}

Table~\ref{tab:full-memory} extends the main-body memory comparison
(Table~\ref{tab:model-memory-comparison}) to all six models.

\begin{table}[!htbp]
  \centering
  \caption{Model-level gradient-computation peak VRAM (GB) across three GPUs,
    all six models.  Same setup as Table~\ref{tab:model-memory-comparison}.
    Values from \texttt{peak\_vram\_mb}.}
  \label{tab:full-memory}
  \small
  \begin{tabular}{llccc}
    \toprule
    \rowcolor{tableheader} \textbf{Model} & \textbf{Method}
      & \textbf{RTX 6000 PRO} & \textbf{H200} & \textbf{B200} \\
    \midrule
    \multirow{4}{*}{Qwen2.5-VL-32B}
      & Eager & OOM & 99.2 & 99.2 \\
      & Fused & OOM & 98.4 & 98.4 \\
      & Dense (B@A) & OOM & 100.6 & 100.6 \\
      & PEFT & OOM & 103.5 & 103.5 \\
    \midrule
    \multirow{4}{*}{Qwen3-VL-32B}
      & Eager & OOM & 99.2 & 99.1 \\
      & Fused & OOM & 98.4 & 98.4 \\
      & Dense (B@A) & OOM & 100.5 & 100.4 \\
      & PEFT & OOM & 103.0 & 102.9 \\
    \midrule
    \multirow{4}{*}{Qwen3.5-27B}
      & Eager & 83.4 & 83.5 & 83.4 \\
      & Fused & 83.3 & 83.4 & 83.3 \\
      & Dense (B@A) & 84.3 & 84.4 & 84.3 \\
      & PEFT & 85.4 & 85.5 & 85.4 \\
    \midrule
    \multirow{4}{*}{Gemma3-27B}
      & Eager & 83.2 & 83.2 & 83.1 \\
      & Fused & 82.9 & 83.0 & 82.9 \\
      & Dense (B@A) & 84.3 & 84.4 & 84.3 \\
      & PEFT & 86.1 & 86.1 & 86.1 \\
    \midrule
    \multirow{4}{*}{Mistral-Sm-24B}
      & Eager & 71.6 & 71.7 & 71.6 \\
      & Fused & 70.6 & 70.7 & 70.6 \\
      & Dense (B@A) & 73.3 & 73.4 & 73.3 \\
      & PEFT & 77.3 & 77.4 & 77.3 \\
    \midrule
    \multirow{4}{*}{Qwen3-VL-8B}
      & Eager & 29.8 & 29.9 & 29.8 \\
      & Fused & 29.5 & 29.5 & 29.5 \\
      & Dense (B@A) & 30.1 & 30.2 & 30.1 \\
      & PEFT & 30.7 & 30.8 & 30.7 \\
    \bottomrule
  \end{tabular}
\end{table}

\section{Single-Layer E2E Decomposition}
\label{app:synthetic-e2e}

The following figures show single-layer end-to-end (E2E) speedup, which
isolates the per-layer overhead but does \emph{not} predict model-level
speedup.  Compose gains compound across $\sim$500 DoRA modules in a real
model, while per-layer backward overhead is amortized, so single-layer E2E
can understate the model-level benefit.

\begin{figure}[p]
  \centering
  \includegraphics[width=\linewidth]{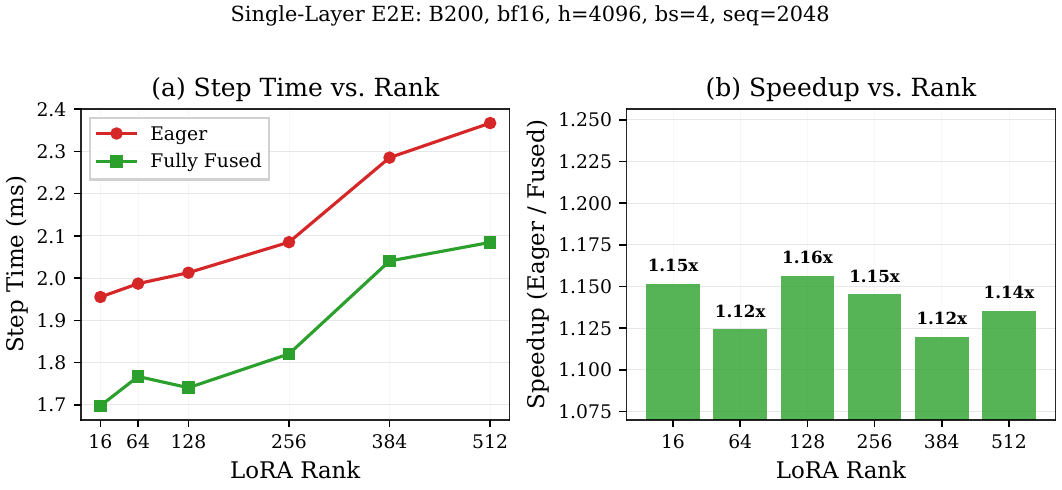}
  \caption{Single-layer E2E overhead decomposition (B200, bf16, $d=4096$,
    bs=4, seq=2048).  Single-layer E2E does not predict model-level speedup:
    compose gains compound across $\sim$500 DoRA modules while per-layer
    backward overhead is amortized.}
  \label{fig:e2e}
\end{figure}

\begin{figure}[p]
  \centering
  \includegraphics[width=0.9\linewidth]{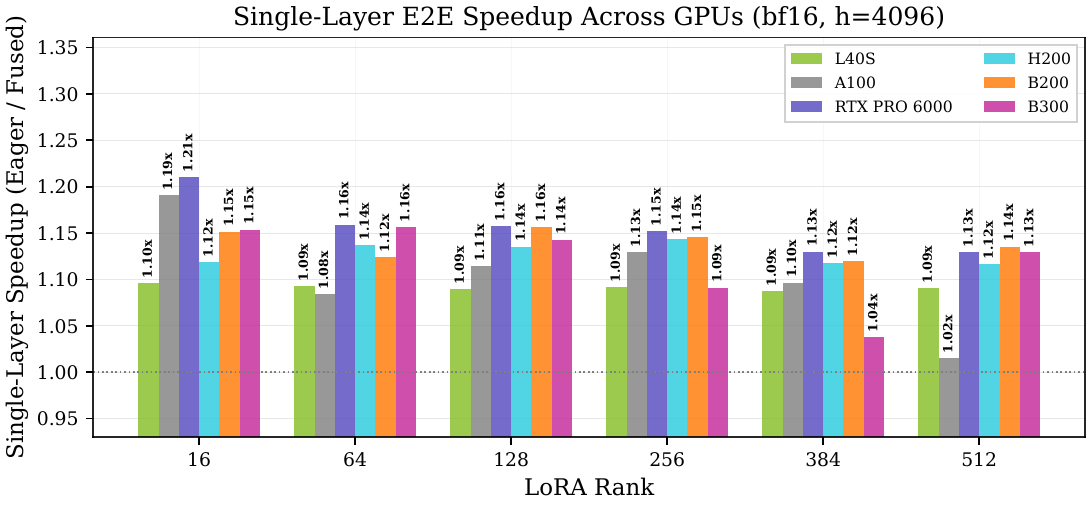}
  \caption{Single-layer E2E speedup (eager/fused) across six GPUs and ranks
    (bf16, $d=4096$, bs=4, seq=2048).  All GPUs show consistent improvement.}
  \label{fig:cross-gpu}
\end{figure}

\begin{figure}[p]
  \centering
  \includegraphics[width=0.85\linewidth]{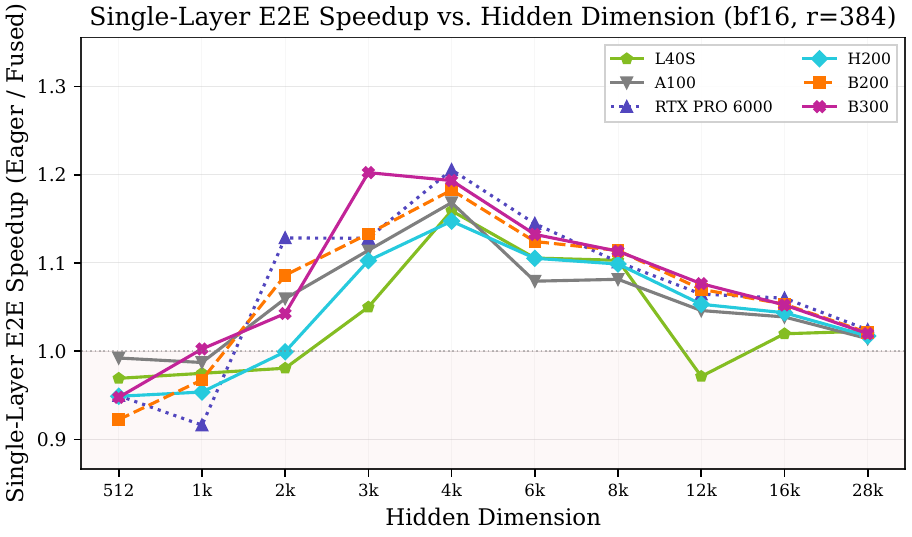}
  \caption{Single-layer E2E speedup vs.\ hidden dimension (bf16, $r=384$,
    six GPUs).  The benefit peaks at $h=3072$--$4096$, corresponding to
    common LLM sizes.}
  \label{fig:hidden-sweep}
\end{figure}

\paragraph{fp32 microbenchmark summary.}
Table~\ref{tab:summary-fp32} provides the fp32 rows omitted from the
main-body summary (Table~\ref{tab:summary}).  Norm memory $3.2$\tms{} in
fp32 reflects the full theoretical benefit, since both paths accumulate in
fp32 and the PEFT baseline also allocates fp32 temporaries.

\begin{table}[!htbp]
  \centering
  \caption{Geometric mean microbenchmark speedups, fp32 (all shapes, 200
    repeats).  Complement to Table~\ref{tab:summary}.}
  \label{tab:summary-fp32}
  \small
  \begin{tabular}{lcccc}
    \toprule
    \rowcolor{tableheader}
      & \textbf{Compose fwd} & \textbf{Backward}
      & \textbf{E2E} & \textbf{Norm mem} \\
    \midrule
    L40S fp32       & 1.53\tms{} & 1.21\tms{} & 1.04\tms{} & 3.2\tms{} \\
    A100 fp32       & 1.64\tms{} & 1.20\tms{} & 1.02\tms{} & 3.2\tms{} \\
    RTX 6000 PRO fp32 & 1.97\tms{} & 1.27\tms{} & 1.06\tms{} & 3.2\tms{} \\
    H200 fp32       & 1.84\tms{} & 1.13\tms{} & 1.03\tms{} & 3.2\tms{} \\
    B200 fp32       & 2.35\tms{} & 1.26\tms{} & 1.03\tms{} & 3.2\tms{} \\
    B300 fp32       & 2.23\tms{} & 1.21\tms{} & 1.03\tms{} & 3.2\tms{} \\
    \bottomrule
  \end{tabular}
\end{table}

\section{Framework Survey}
\label{app:framework-survey}

Table~\ref{tab:framework-survey} summarizes the DoRA norm implementation across
five major fine-tuning frameworks as of February~2026.  We manually inspected
the DoRA-related source code in each framework's main branch at the specified
commits/versions, searching for norm computation implementations.  Paths are
shown relative to each framework's source root for readability.  All use the
same dense-materialization algorithm; none offer a memory-efficient alternative.

\begin{table}[!htbp]
  \centering
  \caption{DoRA norm implementation in major fine-tuning frameworks (February~2026).}
  \label{tab:framework-survey}
  \footnotesize
  \setlength{\tabcolsep}{5pt}
  \begin{tabular}{@{}l l p{4.6cm} p{3.2cm}@{}}
    \toprule
    \rowcolor{tableheader} \textbf{Framework} & \textbf{Version} & \textbf{File Path} & \textbf{Pattern} \\
    \midrule
    HF PEFT & \texttt{20a9829} &
      \path{peft/tuners/lora/dora.py} &
      \texttt{torch.eye} \\[2pt]
    torchtune & \texttt{v0.5.0} &
      \path{modules/peft/dora.py} &
      Same algorithm \\[2pt]
    Unsloth & \texttt{2026.3.7} &
      \normalfont Disables custom kernels &
      Falls back to PEFT \\[2pt]
    SWIFT (ms-swift) &
      \texttt{a807cb9} &
      \normalfont Defers to PEFT/Unsloth &
      No custom code \\[2pt]
    LLaMA-Factory & \texttt{v0.9.3} &
      \normalfont Delegates to PEFT &
      No custom code \\[2pt]
    Axolotl & \texttt{v0.6.0} &
      \normalfont Delegates to PEFT &
      No custom code \\
    \bottomrule
  \end{tabular}
\end{table}

\end{document}